\documentclass[sigconf,natbib=true,anonymous=false,authorversion=true]{acmart}

\AtBeginDocument{%
  \providecommand\BibTeX{{%
    \normalfont B\kern-0.5em{\scshape i\kern-0.25em b}\kern-0.8em\TeX}}}

\copyrightyear{2025}
\acmYear{2025}
\setcopyright{cc}
\setcctype{by}


\usepackage{listings}
\usepackage{balance}
\usepackage{color}
\usepackage{xcolor}
\usepackage{pifont}

\usepackage{colortbl}
\usepackage{subcaption}

\usepackage{lipsum}
\usepackage{verbatim}
\usepackage{colortbl}
\usepackage{multirow}
\usepackage{soul}

\usepackage[linesnumbered,ruled,vlined]{algorithm2e}

\usepackage{enumitem}

\newlist{questions}{enumerate}{2}
\setlist[questions,1]{label=\textbf{(Q\arabic*)},ref=\textbf{(Q\arabic*)}}
\setlist[questions,2]{label=(\alph*),ref=\thequestionsi(\alph*)}

\newcommand{\mathcolorbox}[2]{\colorbox{#1}{$\displaystyle #2$}}
\NewDocumentCommand{\codeword}{v}{%
    \texttt{\textcolor{violet}{#1}}%
}

\NewDocumentCommand{\colorcodeword}{vv}{%
    \textbf{\texttt{\textcolor{#1}{#2}}}%
}

\makeatletter
\def\NAT@def@citea{\def\@citea{\NAT@separator}}
\makeatother

\definecolor{lightgray}{gray}{0.95}
\definecolor{lightpeach}{HTML}{FCE5CD}
\definecolor{softpink}{RGB}{245, 220, 225}
\definecolor{softblue}{HTML}{dae8fc}

\begin{document}



\title{A Super Fast K-means for Indexing Vector Embeddings}



\settopmatter{authorsperrow=3}

\author{Leonardo Kuffo}
\affiliation{%
  \institution{CWI}
  \city{Amsterdam}
  \country{The Netherlands}}
\email{lxkr@cwi.nl}

\author{Sven Hepkema}
\affiliation{%
    \institution{Systems Group, ETH Zurich}
    \city{Zurich}
  \country{Switzerland}
}
\email{sven.hepkema@inf.ethz.ch}

\author{Peter Boncz}
\affiliation{%
  \institution{CWI}
  \city{Amsterdam}
  \country{The Netherlands}}
\email{boncz@cwi.nl}



\begin{abstract}

We present \textit{SuperKMeans}: a $k$-means variant designed for clustering collections of high-dimensional vector embeddings. SuperKMeans' clustering is up to 7x faster than FAISS and Scikit-Learn on modern CPUs and up to 4x faster than cuVS on GPUs (Figure 1), while maintaining the quality of the resulting centroids for vector similarity search tasks. SuperKMeans acceleration comes from reducing data-access and compute overhead by reliably and efficiently pruning dimensions that are not needed to assign a vector to a centroid. Furthermore, we present \textit{Early Termination by Recall}, a novel mechanism that early-terminates $k$-means when the quality of the centroids for retrieval tasks stops improving across iterations. In practice, this further reduces runtimes without compromising retrieval quality. We open-source our implementation at \url{https://github.com/cwida/SuperKMeans}. 



\end{abstract}

\maketitle




\section{Introduction}\label{sec:intro}

The $k$-means algorithm is a classic clustering technique that organizes multi-dimensional points into $k$ clusters~\cite{kmeans}. Recently, $k$-means has been utilized in \textit{approximate vector similarity search} (VSS) to index large collections of high-dimensional vectors~\cite{quake, scannwhitepaper, lorann, rabitq, anisotropic, rabitqext}. VSS consists of finding the vectors in a collection that are the most similar to a given query vector, based on a distance or similarity metric. Once a vector collection has been indexed using $k$-means, the resulting centroids serve as entry points to guide queries to the clusters most likely to contain their nearest neighbors. Thus, reducing search latency by avoiding accessing the majority of the collection, albeit only providing \textit{approximate} answers. This indexing method, commonly known as Inverted-Files (IVF)~\cite{pqivf, faisspaper}, is one of the most widely used indexes in vector systems \cite{faisspaper, micronn, vectordbs, milvus, quake}.

However, in modern vector search applications (e.g., information retrieval engines, recommendation systems, RAG systems), vectors have dimensions in the hundreds and thousands. Additionally, the number of clusters ($k$) needed for vector indexes can be orders of magnitude higher than the typical $k$ in other applications. The latter makes $k$-means heavily compute-bound. As a result, the time to index a collection is a major pain point among users of vector systems. Despite this, the implementation of $k$-means in modern vector systems like FAISS~\cite{faisspaper} has not evolved beyond vanilla $k$-means (i.e., Lloyd's $k$-means) with tuned hyperparameters. This is because Lloyd's $k$-means can leverage highly optimized General Matrix Multiplication (GEMM) routines to accelerate the pairwise distance calculation between data points and cluster centroids---the main bottleneck of $k$-means. In contrast, newer variants of $k$-means cannot take advantage of GEMM routines and lack comprehensive evaluation on vector embedding datasets~\cite{zhakubayev2024beta, zhakubayev2022clustering, marigold, hamerly2014accelerating}. 

\begin{figure}[t!]
\centering
\includegraphics[width=1.0\columnwidth]{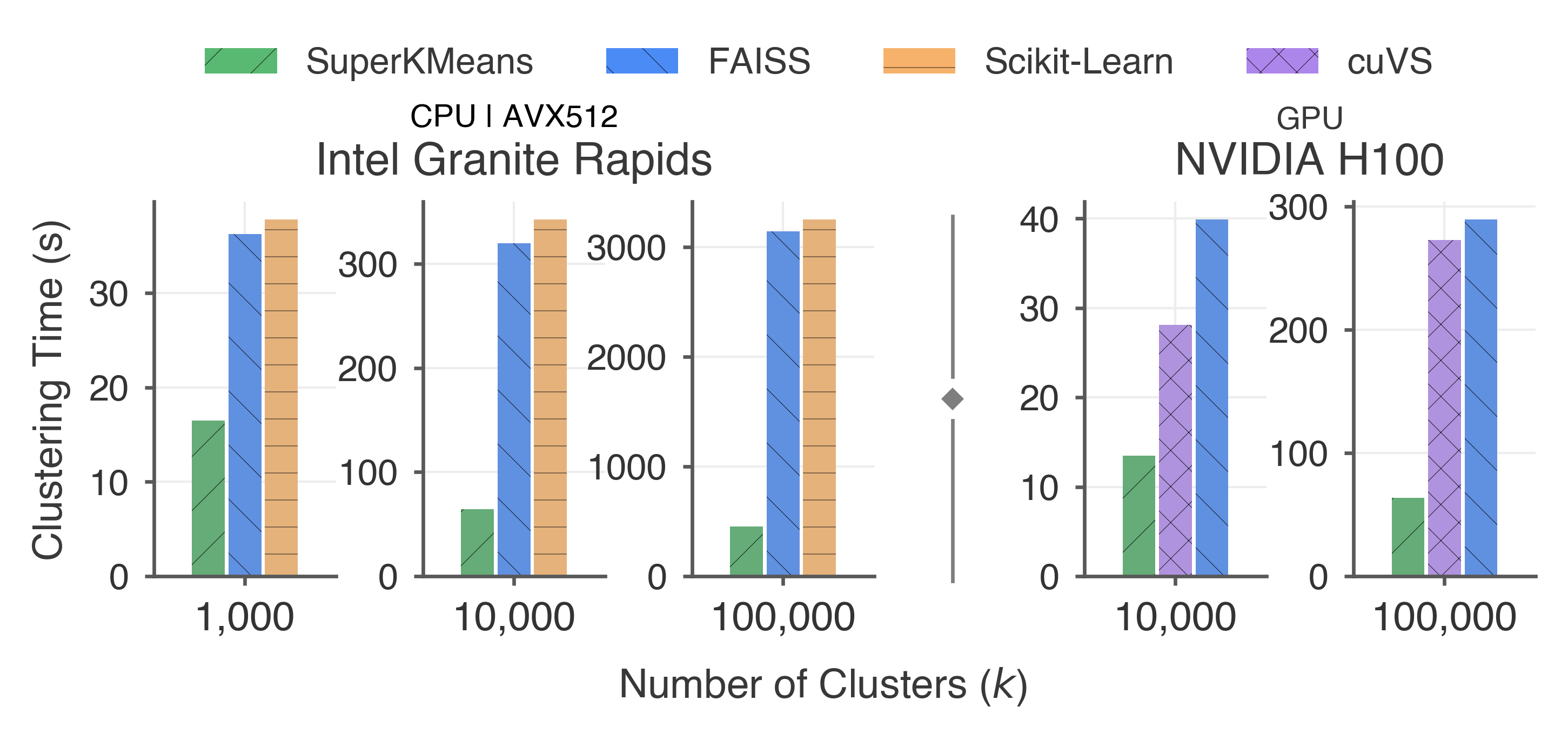}
\vspace*{-9mm}
\caption{Clustering performance for all competitors in the OpenAI dataset with 1M vector embeddings of 1536 dimensions. SuperKMeans performs exceptionally well as
the number of clusters increases (up to 7x faster on CPU and 4x on GPU), making it attractive for indexing large embedding collections. The evaluation framework is presented in Section~\ref{sec:eval}.
}
\label{fig:opening}
\vspace*{-4mm}
\end{figure}

In this work, we present SuperKMeans, a \underline{super}-fast \underline{k-means} variant tailored for building partition-based indexes in collections of high-dimensional vectors. SuperKMeans drastically reduces data accesses and compute overhead in the core loop of $k$-means by pruning dimensions that are not needed to determine whether a vector will be assigned to a centroid. SuperKMeans' novelty lies in carefully interleaving GEMM routines and pruning kernels during the pairwise distance calculation phase, a feat not previously accomplished by other $k$-means variants. Efficient pruning kernels are achieved by using Adaptive Sampling (ADSampling)~\cite{adsampling} and by keeping centroids in memory using the PDX layout~\cite{pdx}. 
As a result, \textbf{SuperKMeans clusters 1M 1536-dimensional vectors 7x faster than FAISS on CPUs and 4x faster than cuVS on GPUs}, while producing clusters that maintain equivalent vector retrieval quality. 
Furthermore, we combine SuperKMeans' acceleration with hierarchical $k$-means~\cite{micronn, turbopuffer, kmeanselastic, jin2026curator, spfresh}. This mix achieves unprecedented performance, indexing 10M 1024-dimensional vector embeddings 250x faster than FAISS.

Finally, we introduce a novel early termination mechanism named \textbf{Early Termination by Recall} (ETR). ETR monitors the \textit{recall} that the current centroids would yield if used to conduct a VSS with either user-supplied queries or queries sampled from the data. An early termination of $k$-means occurs when the quality of VSS does not improve between iterations. This mechanism shortens index construction time and alleviates the need for users to guess how many iterations are needed to achieve optimal retrieval quality. Importantly, ETR is not tied to SuperKMeans, as it can be integrated into any IVF indexing pipeline.

Our main contributions are:
\begin{itemize}
    \item The design of SuperKMeans, a $k$-means variant for indexing collections of high-dimensional vector embeddings.
    \item A comprehensive evaluation of SuperKMeans when clustering various vector embedding collections against Scikit-Learn~\cite{scikit} and FAISS~\cite{faisscode} on different CPUs (x86: Intel Granite Rapids, Zen 5, and Zen 3, ARM: Apple M4 and Graviton 4), demonstrating its superiority and portability.
    \item The design of SuperKMeans on GPUs, alongside an evaluation against FAISS and cuVS~\cite{cuvs} presented in Section~\ref{sec:gpu}.
    \item The insights that, in vector embedding datasets: i) just 5 to 10 iterations of $k$-means are enough to achieve optimal retrieval quality, ii) using 20-30\% of the data points is sufficient to obtain high-quality centroids, and iii) that using $k$-means++ initialization is detrimental for index quality.
    \item An open-source implementation of SuperKMeans available at: \url{https://github.com/cwida/SuperKMeans}
\end{itemize}

\section{Preliminaries}\label{sec:preliminaries}

\begin{table}[]
\centering
\caption{Terminology}
\vspace*{-4mm}
\label{tab:terminology}
\resizebox{0.8\linewidth}{!}{%
\begin{tabular}{ll}
\hline
\textbf{Symbol} & \textbf{Definition}                                           \\ \hline
\textbf{$X$}      & Matrix of data vectors                                                 \\
\textbf{$x$}      & A vector from $X$                                                 \\
\textbf{$N$}      & Number of data vectors                                        \\
\textbf{$Y$}      & Matrix of clusters' centroids                                            \\
\textbf{$y$}      & A centroid from $Y$                                            \\
\textbf{$k$}      & Number of clusters to compute with $k$-means                      \\
\textbf{$d$}      & Dimensionality of X and Y                                     \\
\textbf{$d'$}     & Cutoff dimension to split X (around 12\% of $d$)                \\
\textbf{$X', Y'$}     & The first $d'$ dimensions of the matrix $X$, $Y$    \\
\textbf{$x', y'$}     & The first $d'$ dimensions of the vector $x$, $y$     \\
\textbf{$d''$}     & Trailing $d - d'$ dimensions                 \\
\hline
\end{tabular}
\vspace*{-6mm}
}
\end{table}

Table~\ref{tab:terminology} presents the terminology used in this study. 

\subsection{K-means}
$K$-means is a half-century-old algorithm for grouping multidimensional points into $k$ clusters~\cite{kmeans}. In brief, the algorithm follows these steps: 1) Generate $k$ initial centroids. 2) Assign each data point $x$ to its \textit{closest} centroid $y$. 3) Update centroids $Y$ based on the points assigned to them. Afterwards, steps 2 and 3 (i.e., the \textit{core loop} of $k$-means) are repeated until a \textit{termination condition} is met. The vanilla version of $k$-means is known as Lloyd's algorithm~\cite{kmeans}. Over the years, numerous variations have been proposed for each step of the algorithm~\cite{marigold, surveygraph, ding2015yinyang, hamerly2014accelerating, elkan2003using, zhakubayev2024using, zhakubayev2024beta}. However, we have observed that these variations often fall short in one or more critical aspects: efficient vectorization (SIMD), CPU/GPU-friendly algorithms, evaluation across CPU microarchitectures, comprehensive evaluation in high-dimensional embeddings datasets ($d > 512$) using a high number of clusters ($k > 10000$), and comparison against SOTA systems. As a result, the vanilla Lloyd's algorithm is the de facto alternative in most clustering libraries for vector embeddings~\cite{faisspaper}.

In this section, we discuss the most important aspects of the $k$-means algorithm. Note that we will not cover other clustering algorithms, such as DBSCAN~\cite{dbscan} and HDBSCAN~\cite{hdbscan}, as they are not commonly used for building vector search indexes because they form clusters based on \textit{density} rather than \textit{proximity} to centroids. Also, we shifted our focus away from methods that use auxiliary data structures for clustering (e.g., seeded search graphs~\cite{usearch, seededsearchgraphs}, SA-trees~\cite{satrees}, cover trees~\cite{covertrees}), as the costs associated with building these structures often exceed those of $k$-means.

\vspace*{3mm}\noindent{\bf STEP 1. Centroids Initialization: } Typically, centroids are initialized by randomly sampling $k$ vectors from $X$. This is also known in the literature as the Forgy initialization~\cite{forgy1965cluster}. A more sophisticated initialization, known as $k$-means++~\cite{kmeansplusplus}, begins by randomly picking the first centroid from $X$. Subsequent centroids are then iteratively selected with a probability proportional to their distance to their nearest existing centroid. In other words, vectors that are further away from the current set of centroids have a higher likelihood to be the next chosen centroid. This initialization has been shown to improve clustering quality while also leading to faster \textit{convergence} (i.e., centroids become stable earlier in the core loop). However, the complexity of this initialization can negate its benefits, particularly in high-dimensional vectors. In such scenarios, the curse of dimensionality makes the boundaries of clusters not well-defined~\cite{franti2019much, satrees, whenisnnmeaningful, meaningfulcurse}. As a result, a random initialization of centroids is as good as doing an initialization with $k$-means++ or other initialization techniques~\cite{bachem2016fast, franti2019much}. These observations are validated by our experiments in Subsection~\ref{sec:eval:kmeans++}.

\vspace*{3mm}\noindent{\bf STEP 2. Determining Assignments: } Assignments are based on a \textit{distance metric} that determines which $x$ is closer to which $y$. The L2 Euclidean distance is the most commonly used, followed by angular metrics such as the inner product. This step is the main bottleneck of $k$-means, as it involves computing pairwise distances between every $x$ in $X$ and every $y$ in $Y$. The most efficient way to do this is through a General Matrix Multiplication (GEMM) between the data points and centroids ($\mathcolorbox{lightpeach}{X \boldsymbol{\cdot} Y}$). Elkan's variant of $k$-means tries to avoid unnecessary distance calculations by applying the triangle inequality to determine whether a distance calculation can be skipped~\cite{elkan2003using}. However, in practice, Elkan's method and its variants (e.g., Hamerly's~\cite{hamerly2014accelerating}), often proves ineffective with high-dimensional vectors due to the curse of dimensionality~\cite{elkan2003using, marigold}, sometimes resulting in worse performance than the vanilla $k$-means algorithm. On top of that, GEMM routines outperform approaches that skip distance calculations, thanks to their optimized data access patterns, batch processing capabilities, vectorization (SIMD), cache efficiency, and efficient multi-threading. These observations are validated by our experiments in Subsection~\ref{sec:eval:elkans}

\vspace*{3mm}\noindent{\bf STEP 3. Updating Centroids: } Centroids are updated by computing the mean of every vector $x$ assigned to them. However, centroids may end up with zero assignments. In such instances, it is beneficial to split large clusters to achieve a more balanced distribution of points per cluster. Another strategy is to replace the centroids of empty clusters with the points that are the furthest from their assigned centroid. Notably, balanced clusters are a desirable characteristic of partition-based indexes used for VSS~\cite{quake, lorann}. 

\noindent{\bf Termination Conditions and Final Assignments: } The core loop terminates after a predetermined number of iterations. However, most $k$-means variants implement an early-exit mechanism when centroids or assignments remain unchanged (or change very little) between iterations. The latter is known as \textit{convergence}. Once the core loop finishes, every $x$ must be assigned to its nearest centroid using the final centroids obtained from the core loop. This is equivalent to performing STEP 2 one last time.
 
\subsection{Keys for the Performance of k-means}
Despite the plethora of research done towards more efficient $k$-means variants~\cite{hamerly2014accelerating, marigold, kmeanssurvey, ding2015yinyang, zhakubayev2024beta, zhakubayev2024using, elkan2003using}, only a handful of open-source implementations achieve SOTA speeds in practice when working with large vector collections. Notably, both FAISS~\cite{faisscode} and Scikit-Learn~\cite{scikit} stand out in this regard. This is because achieving SOTA performance requires not only clever algorithmic work but also co-design with modern hardware. One example is Elkan's $k$-means variant, which achieves theoretical speedups through pruning but, in practice, is slower than a complete $X \boldsymbol{\cdot} Y$ GEMM for distance calculations. We show this in Subsection~\ref{sec:eval:elkans}. This is because $k$-means variants overlook other factors that affect computational efficiency, such as data access patterns, cache utilization, and code vectorization (SIMD). On top of that, it is challenging to compete with GEMM routines in \textit{BLAS libraries}, given their decades of optimization. BLAS libraries are accelerators for linear algebra operations tailored for modern CPUs. For our purposes, we are only interested in the \codeword{sgemm} routine inside BLAS that does a \underline{s}ingle precision (\codeword{float32}) \underline{ge}neral \underline{m}atrix \underline{m}ultiply. Some notable examples of BLAS implementations include IntelMKL for Intel CPUs, AOCL-BLIS for AMD CPUs, Apple Accelerate for Apple Silicon chips, and OpenBLAS, which can be compiled for various CPU microarchitectures. Apple Accelerate uses AMX, a co-processor in Apple Silicon chips specialized in matrix multiplication operations~\cite{amxpaper, corsix_amx}.

\subsection{Vector Indexing and k-means}\label{sec:kmeans-vectorindex}

Recently, $k$-means has received unprecedented attention from the information retrieval community~\cite{lorann, rabitqext, rabitq, faisspaper, spann, quake, li2025saq, anisotropic, micronn, flashkmeans}. This is because it is a relatively fast and powerful algorithm for building \textit{approximate indexes} of high-dimensional vectors. Approximate indexes are data structures that guide a query vector $q$ to the most suitable place of the $d$-dimensional space in which the query \textit{may} find its closest neighbors. As a result, query speed is accelerated by orders of magnitude since the distance metric is only evaluated between $q$ and a few vectors $x \in X$. Approximate indexes lead to \textit{approximate answers}, which are generally acceptable in modern vector-based applications. The quality of these retrieved answers is typically measured using the \textit{recall} metric, which quantifies the proportion of retrieved vectors that are truly the closest to the query. These \textit{actual} closest points are found offline via a brute-force search that calculates the distance from the query $q$ to every $x \in X$. 

\begin{figure}[t!]
\centering
\includegraphics[width=0.7\columnwidth]{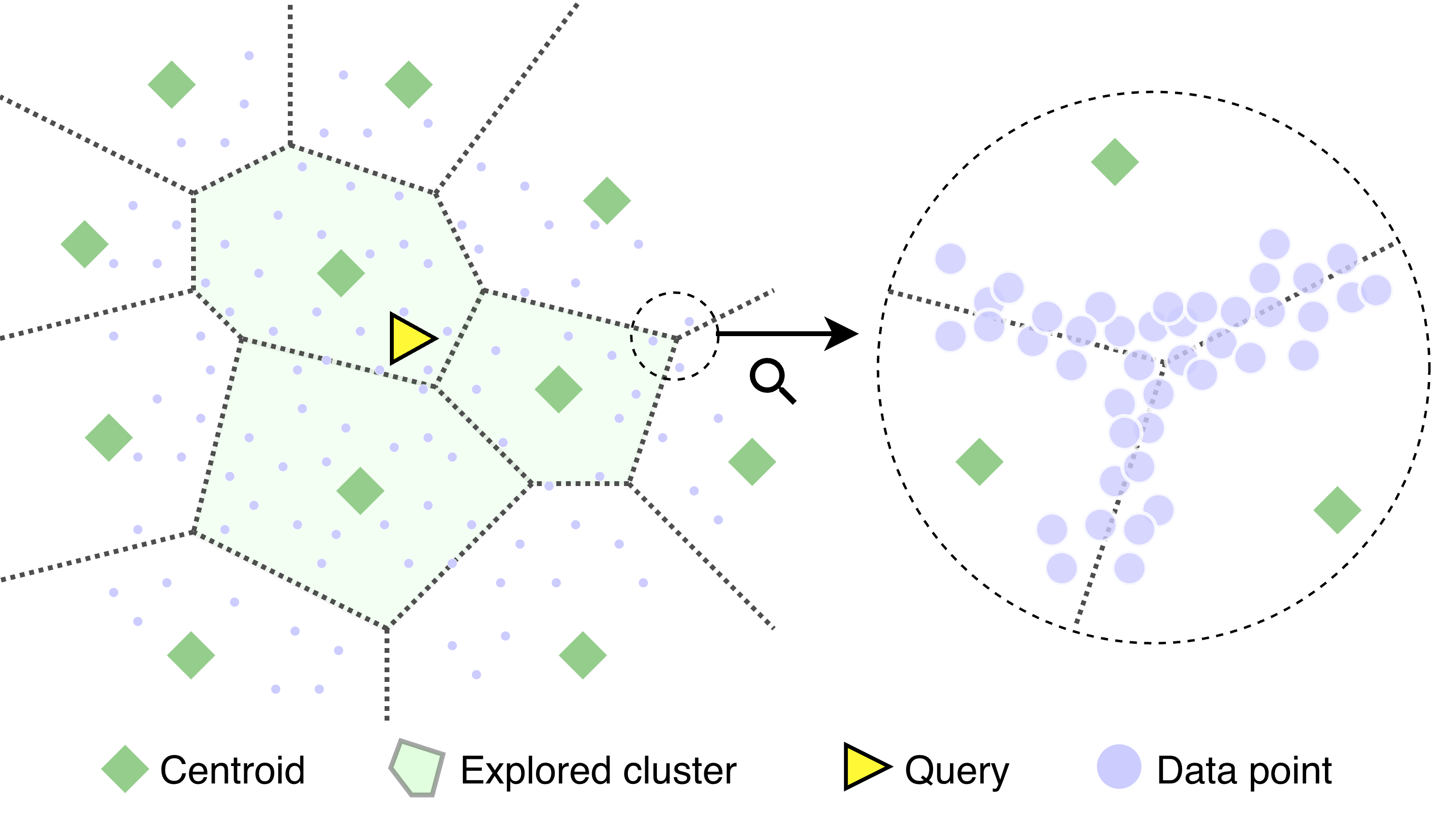}
\vspace*{-3.0mm}
\caption{Example of an IVF index search where only the points in the clusters closest to the query are explored. The clusters are defined via k-means. In high-dimensional vectors, the boundaries of these clusters are not well-defined. This fuzziness makes k-means++ initialization ineffective.}
\vspace*{-4.0mm}
\label{fig:ivf}
\end{figure}

Approximate indexes can be categorized into three types: graph-based~\cite{hnsw, hcnng, surveygraph, patel2024acorn, anisotropic}, partition-based~\cite{quake, micronn, pqivf, surveylsh}, and hybrids, which combine both graph- and partition-based approaches~\cite{spann, ngt}. Partition-based indices, like Inverted-Files (IVF)~\cite{pqivf}, generally group vectors into clusters using $k$-means~\cite{faisspaper, lorann, scannwhitepaper}. During query time, the distance metric is first evaluated between the query $q$ and the centroids $Y$. Then, the vectors inside the nearest clusters are chosen for evaluation (left of Figure ~\ref{fig:ivf}). A higher number of clusters can be explored to trade off search speed for more recall~\cite{faisspaper, milvus}. The number of centroids to use is typically set to the order of $\sqrt{N}$~\cite{faisspaper, milvus}, which leads to higher $k$ values compared to other applications of $k$-means~\cite{marigold, kmeansplusplus}.  Notably, IVF performs modestly well across most datasets~\cite{annbench, candy, vibe} and scales better than graph indexes, which have higher memory requirements and longer construction times~\cite{faisspaper, scannwhitepaper, quake, lorann}. This makes IVF an attractive indexing algorithm for large collections and distributed or cloud environments~\cite{turbopuffer, bann, cloudvss}.

\subsection{Lack of Specialization Won the Race?}\label{sec:kmeans-lack}

There have been no variants of $k$-means specifically designed for indexing vector embeddings. Most state-of-the-art libraries and vector systems use a vanilla $k$-means implementation~\cite{rabitqlibrary, lorann, faisspaper, milvus, weaviate, lance, cuvs}. Among these, the open-source FAISS library is considered state-of-the-art for clustering vector embeddings, both on CPUs and GPUs, thanks to its integration with cuVS~\cite{cuvsblog}. Surprisingly, a key aspect of this implementation is its default hyperparameters, which are optimized for vector indexing. First, FAISS limits the number of iterations to 25 for general-purpose $k$-means and 10 for constructing an IVF index, whereas other implementations, such as Scikit-Learn's~\cite{scikit}, set the number of iterations in the hundreds~\cite{hamerly2014accelerating}. Second, FAISS subsamples the data used for clustering, up to a maximum of $256 * k$ points, which effectively reduces the runtime of the algorithm. Finally, FAISS employs random centroid initialization by default. Thus, avoiding more complex initialization methods such as $k$-means++~\cite{kmeansplusplus} or using core loop variants such as Elkan's~\cite{elkan2003using} or Hamerly's~\cite{hamerly2010making}.

As previously mentioned, this lack of specialization stems from the fact that, in high-dimensional embeddings, the curse of dimensionality significantly impacts the performance of most $k$-means variants. This phenomenon leads to densely populated boundaries of the Voronoi cells. In other words, in most vector embedding datasets \textbf{clusters are not well defined}~\cite{spann, xu2025scalable, satrees, meaningfulcurse}. Figure~\ref{fig:ivf} illustrates this phenomenon. Consequently, a random initialization of clusters is as effective as one generated by $k$-means++, and attempts to prune with the triangle inequality offer little benefits~\cite{elkan2003using, ding2015yinyang, kmeanselastic}. Another consequence of poorly defined clusters is that the quality of the centroids to satisfy vector retrieval tasks converges after just a few iterations~\cite{kmeanselastic}. These observations are validated in our experiments in Subsections~\ref{sec:eval:etr}, ~\ref{sec:eval:kmeans++} and ~\ref{sec:eval:elkans}.

\section{SuperKMeans: The Algorithm}\label{sec:superkmeans}

SuperKMeans is a variant of the $k$-means algorithm designed for clustering collections of high-dimensional vector embeddings. The secret sauce of SuperKMeans is reducing data accesses and compute overhead by avoiding unnecessary distance calculations for centroids unlikely to be a vector assignment. To accomplish this \textit{efficiently}, SuperKMeans divides an iteration of the $k$-means core loop into two phases: \textbf{\colorbox{lightpeach}{GEMM}} and \textbf{\colorbox{softblue}{PRUNING}}. The \colorbox{lightpeach}{GEMM} phase calculates partial distances only with the $d'$ front dimensions of the data vectors using GEMM routines. Then, in the \colorbox{softblue}{PRUNING} phase, these partial distances are evaluated to determine which centroids are unlikely to be assigned to a vector, leading to the pruning of their trailing $d''$ dimensions. This pruning process incurs nearly zero loss thanks to a technique known as Adaptive Sampling (ADSampling), which has been used to accelerate VSS algorithms~\cite{adsampling, pdx, rabitq}. ADSampling offers much higher, more effective pruning capabilities than the triangle inequality~\cite{pdx, adsampling, rabitq}. Remarkably, in most datasets, more than 97\% of centroids are pruned at the start of the \colorbox{softblue}{PRUNING} phase. Furthermore, SuperKMeans incorporates \textit{Early Termination by Recall}. This novel early-termination mechanism stops the algorithm when the centroids no longer improve the \textit{recall} they yield when answering VSS queries across iterations. A referential pseudocode is presented in Algorithm~\ref{algo:pseudocode}. 


\begin{figure}[t!]
\centering
\includegraphics[width=0.88\linewidth]{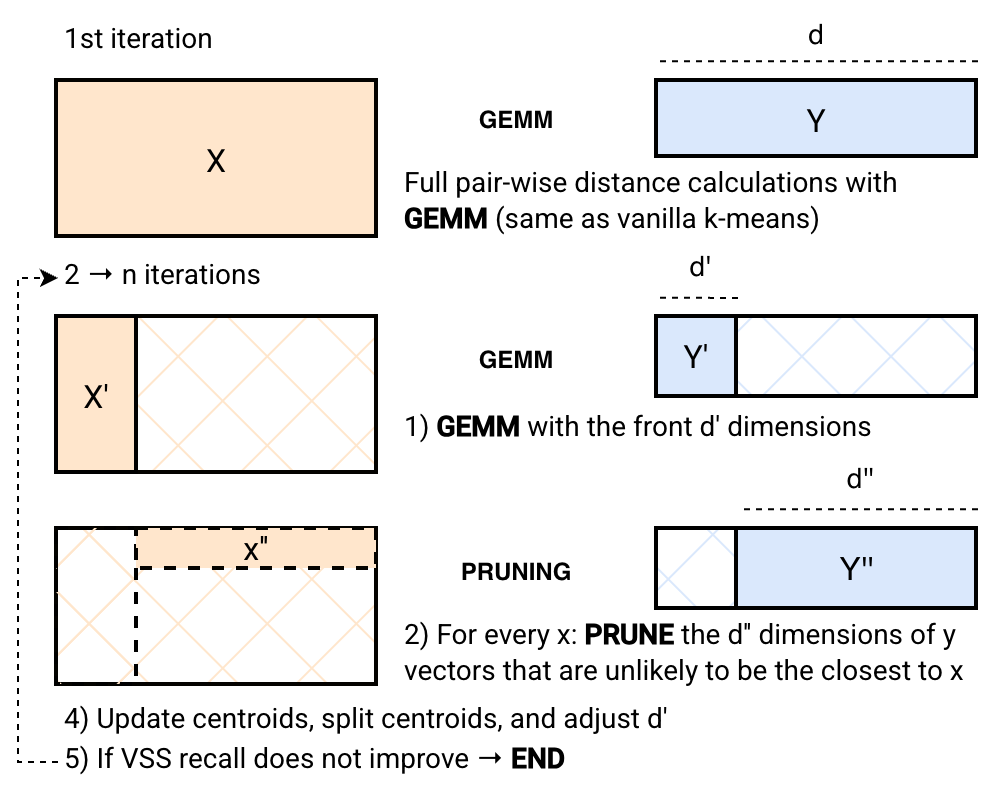}
\vspace*{-5.0mm}
\caption{SuperKMeans divides an iteration into two phases: i) GEMM: Computes partial distances with the front d' dimensions, and ii) PRUNING: prunes the d'' dimensions of the centroids unlikely to be the assignment of a vector. }
\vspace*{-2.0mm}
\label{fig:coreloop}
\end{figure}

\begin{algorithm}[t!]

\definecolor{lightgray}{gray}{0.6}

\footnotesize

\SetAlgoNoEnd

\setlength{\algomargin}{1em}       
\SetAlgoVlined                     
\SetAlgoNlRelativeSize{-2} 
\SetAlgoInsideSkip{0pt}
\SetAlCapFnt{\small} 
\SetKwComment{tcp}{}{}
\SetAlgoSkip{0pt}


\caption{Pseudocode of SuperKMeans}
\label{algo:pseudocode}
\KwIn{$X \in \mathbb{R}^{N \times d}$, $k$, $n\_iters$}
\KwOut{$Y \in \mathbb{R}^{k \times d}$}

$R \gets \textsc{GenerateRandomRotationMatrix}(d)$ \tcp{ \textcolor{lightgray}{\textnormal{\footnotesize // Random rotation matrix}}}
$\tilde{X} \gets X \cdot R$ \tcp{ \textcolor{lightgray}{\textnormal{\footnotesize // Rotate data}}}
$\tilde{Y} \gets \textsc{InitCentroids}(\tilde{X}, k)$ \tcp{ \textcolor{lightgray}{\textnormal{\footnotesize // Randomly sample centroids}}}
$d' \gets \lfloor d * 12.5\% \rfloor$

\tcp{ \textcolor{gray}{\textnormal{\small --- Iteration 1: Full GEMM ---}}}

$distances \gets -2 \mathcolorbox{lightpeach}{\tilde{X} \cdot \tilde{Y}} + \|\tilde{X}\|^2 + \|\tilde{Y}\|^2$\;
\For{$i = 1$ \KwTo $N$}{
    $assignments[i] \gets \arg\min_j distances[i,j]$\;
}
$\tilde{Y} \gets \textsc{UpdateCentroidsAndSplitEmpty}(\tilde{X}, assignments, k)$\;

\BlankLine

\tcp{ \textcolor{gray}{\textnormal{\small --- Rest of Iterations: Partial GEMM + PRUNING}}}

\For{$t = 2$ \KwTo $n\_iters$}{ 

    $\tilde{Y} \gets \textsc{PDXification}(\tilde{Y})$\;

    $distances \gets -2 \space \mathcolorbox{lightpeach}{\tilde{X}_{1:d'} \cdot \tilde{Y}_{1:d'}} + \|\tilde{X}_{1:d'}\|^2 + \|\tilde{Y}_{1:d'}\|^2$ \tcp{ \textcolor{lightgray}{\textnormal{\footnotesize // GEMM}}}

    \For{each point $\tilde{x}_i$}{
        $\tau \gets \|\tilde{x}_i - \tilde{y}_{assignments[i]}\|^2$ \tcp{ \textcolor{lightgray}{\textnormal{\footnotesize // Initial pruning threshold}}}

        \For{each centroid $\tilde{y}_j$}{

            \If{$likely(distances[i,j] > \theta(d', \tau))$} 
            { 
                \textbf{continue} \tcp{ \textcolor{lightgray}{\textnormal{\footnotesize // ADSampling's pruning}}}
            }

            \For{PDXBlocks $b = d' \dots d$}{
                $distances[i,j] \mathrel{+}= \|\tilde{x}_i[b] - \tilde{y}_j[b]\|^2$\;
                \If{$distances[i,j] > \theta(b, \tau)$}{
                    \textbf{break} \tcp{ \textcolor{lightgray}{\textnormal{\footnotesize // Prune remaining blocks}}}
                }
            }

            \If{$distances[i,j] < \tau$}{
                $assignments[i] \gets j$\;
                $\tau \gets distances[i,j]$\;
            }
        }
    }

    $\tilde{Y} \gets \textsc{UpdateCentroidsAndSplitEmpty}(\tilde{X}, assignments, k)$\;

    $d' \gets \textsc{AdjustD'}()$\;

    \If{VSS recall does not improve}{
        \textbf{break} \tcp{ \textcolor{lightgray}{\textnormal{\footnotesize // ETR}}}
    }
}

\Return{$R^{-1} \cdot \tilde{Y}$} \tcp{ \textcolor{lightgray}{\textnormal{\footnotesize // Return unrotated centroids}}}

\end{algorithm}

\subsection{Preprocessing}

\vspace*{3mm}\noindent{\bf Random Rotation: } To use ADSampling's pruning, we need to perform a random orthogonal rotation of $X$. This operation spreads the variance evenly across all dimensions while preserving both angular and Euclidean distances between vectors. The outcome of this is that any single dimension behaves like a random projection of the original vector, carrying a representative "random slice" of its variance. Thanks to this, one can determine via hypothesis testing when a partial distance is already too \textit{far} to end up being closer than the current \textit{closest} distance. 
We refer readers to the original ADSampling's paper~\cite{adsampling} for a more comprehensive explanation. This random rotation is effectively a GEMM between $X$ and a rotation matrix $R$. $R$ is a square matrix of size $d$. Since the size of $R$ is limited, the total time spent in this preprocessing is negligible compared to the core loop of $k$-means. 

\vspace*{3mm}\noindent{\bf Centroids Initialization:} We randomly sample $k$ vectors from $X$ to use them as initial centroids. As described in subsection~\ref{sec:kmeans-lack}, we want our centroid initialization to be as quick as possible. Therefore, we avoid more sophisticated approaches such as $k$-means++. 

\vspace*{3mm}\noindent{\bf Norms Calculation: } The GEMM between $X$ and $Y$ ($\mathcolorbox{lightpeach}{X \boldsymbol{\cdot} Y}$) essentially computes an inner product. However, we are interested in the L2 distance metric. Fortunately, we can obtain L2 distances from inner products by applying the identity for \textit{L2 distance expansion via inner products}: $||x-y||^2 = ||x||^2 + ||y||^2 - 2*(\mathcolorbox{lightpeach}{X \boldsymbol{\cdot} Y})$. To use this identity, we must  keep the norms of every $x$ and every $y$. However, recall that our distance calculation is divided in two phases: \colorbox{lightpeach}{GEMM} and \colorbox{softblue}{PRUNING}. Let us say that the GEMM phase takes the front $d'$ dimensions. Then, the PRUNING phase takes the trailing $d'' = d - d'$ dimensions of $X$ and $Y$. Thankfully, the L2 distance expansion via inner products is applicable for partial calculations as well, since: $||x-y||^2 = \mathcolorbox{lightpeach}{||x'-y'||^2} + \mathcolorbox{softblue}{||x''-y''||^2}$. To do this, we need two additional buffers that store the partial norms at $d'$ of every $x'$ and every $y'$. For the \colorbox{softblue}{PRUNING} phase, we do not use the identity, as we no longer perform a matrix multiplication to compute distances.


\subsection{Core Loop}

The core loop of SuperKMeans is illustrated in Figure~\ref{fig:coreloop}. 

\vspace*{3mm}\noindent{\bf Batch Processing Principle:} $\mathcolorbox{lightpeach}{X \boldsymbol{\cdot} Y}$ produces an output matrix of size $N * k$. If $N=1M$ and $k=4000$, one would require a buffer of 16 GB. To avoid these large temporary buffers, we implemented a batching mechanism that works as follows: $\mathcolorbox{lightpeach}{X \boldsymbol{\cdot} Y}$ is done in batches using a nested \codeword{for} loop. The outer loop iterates through $X$ in batches of 4096 vectors, and the inner loop iterates through $Y$ in batches of 1024 vectors. That means that our temporary buffer for output distances is only 16 MB (fits in L3 cache). After $X \boldsymbol{\cdot} Y$, we iterate over the 4096 vectors of $X$. For each $x$ vector, we: i) Apply the L2 distance expansion via inner products to get the L2 distances between $x$ and every $y$ in the batch. If this distance is smaller than the current closest distance for that vector, we update its assignment. This batching strategy is also implemented in FAISS~\cite{faisscode}. From here on, when we refer to $X$ and $Y$, we will mean a single batch. Note that  this batching mechanism is not shown in the pseudocode.

\vspace*{3mm}\noindent{\bf 1st iteration:} We compute a complete pair-wise distance calculation between $X$ and $Y$ with \colorbox{lightpeach}{GEMM}. This is the same as what is done in any vanilla $k$-means implementation in all iterations.

\vspace*{3mm}\noindent{\bf 2nd → n iteration:} We do \colorbox{lightpeach}{GEMM} only with the front $d’$ dimensions of $X$ and $Y$ (Algorithm~\ref{algo:pseudocode} Line 11). Then, for every $x'$ in $X'$ and for every $y'$ in $Y'$, we apply the identity in order to get the partial L2 distances at $d’$. Next, we start the \colorbox{softblue}{PRUNING} phase. For every $x$ in $X$, we determine which $y$ centroids cannot be $x$'s closest centroid using ADSampling hypothesis testing. On the remaining $y$ that were not pruned, we explore the following $\Delta d$ dimensions in hopes of discarding the centroid. We keep exploring the remaining $y$ centroids until we reach $d$, in which case $y$ would be discarded or kept as the assigned centroid of $x$. 
Per our experiments, $d'$ is initially set to 12.5\% of $d$ and $\Delta d$ is set to 64.

\vspace*{3mm}\noindent{\bf In between iterations:} \underline{Update Centroids:} We update every centroid by calculating the mean of all vectors assigned to it. We also update their partial norms. \underline{Split Centroids:} If there are empty clusters, we randomly split a cluster into two, giving larger clusters a higher probability of being picked. To split the centroid, we add a small symmetric perturbation to each dimension of the centroid. This process helps maintain the clusters' balanced and is the same splitting process used in FAISS. \underline{Adjust $d'$:} There is a sweet spot for $d'$ that achieves an optimal performance. If $d'$ is too low, the distances resulting from the partial \colorbox{lightpeach}{GEMM} will not be sufficient for \colorbox{softblue}{PRUNING} to avoid enough work. As a result, the SuperKMeans iteration can be slower than doing a full GEMM. 
On the other hand, if $d'$ is too high, \colorbox{lightpeach}{GEMM} will do unnecessary work, thereby not fully exploiting the potential of the two-phase approach. This sweet spot varies across datasets. In subsection~\ref{sec:eval:sweet} we present an in-depth study of the $d'$ parameter and how we adjust it during runtime. 

\subsection{Termination and Final Assignments}
SuperKMeans implements three termination conditions. The first one is reaching a maximum number of iterations. The second one is reaching convergence, and the third one is \textit{Early Termination by Recall (ETR)}. When ETR is enabled, SuperKMeans monitors the recall that the centroids from the current iteration would achieve if used to perform a VSS on a set of user-supplied queries or sampled from $X$. If the recall does not improve by more than a specified threshold over the past two iterations, the algorithm terminates. This patience threshold of two iterations is used to prevent the algorithm from settling into a local optimum. In subsection~\ref{sec:eval:etr}, we explore this threshold in detail. Note that measuring recall requires \textit{ground truth} data. If the user does not provide this, we compute it at runtime. In Subsection~\ref{sec:eval:microbench}, we show that computing a ground truth incurs minimal overhead compared to the core loop. Upon termination, we accelerate the final assignment phase of the algorithm by employing our \colorbox{lightpeach}{GEMM} + \colorbox{softblue}{PRUNING} approach.

\subsection{Key optimizations}\label{sec:alg:opt}
The \colorbox{softblue}{PRUNING} phase must be as efficient as possible to beat a full GEMM at each iteration. To this end, we introduce two novel optimizations that are crucial for high performance: i) Progressive pruning and ii) using an initial threshold for the PRUNING phase. In Subsection~\ref{sec:eval:ablation}, we conduct an ablation study to show how each optimization contributes to overall performance. 

\vspace*{3mm}\noindent{\bf Progressive Pruning:} Previous $k$-means variants, such as Elkan's, perform binary pruning. That means that when a centroid cannot be discarded by using the triangle inequality, it is necessary to compute the full distance between the vector and the centroid. In contrast, SuperKMeans adopts a progressive pruning approach (Algorithm~\ref{algo:pseudocode} Lines 17-20). If a centroid remains viable for assignment after the partial GEMM, we keep trying to prune the centroid every 64 dimensions. As more dimensions are explored, the resolution of the distance metric improves, thereby increasing the likelihood of pruning the centroid~\cite{adsampling}. However, naively skipping distance calculations negatively affects performance due to inefficient access patterns and cache utilization~\cite{pdx, tribase}. To maximize the performance of pruning, we store the $d''$ dimensions in a block-column-major order of 64 dimensions. This layout enables efficient pruning of dimensions thanks to its optimized data-access patterns~\cite{pdx,pdx2}. We call this process \textbf{PDXification}
as this layout is inspired by the PDX layout~\cite{pdx}. 
Note that, since $d''$ is not always a multiple of 64, we allow for a tailing block of less than 64 dimensions.  



\vspace*{3mm}\noindent{\bf Initial threshold for \colorbox{softblue}{PRUNING}:} At the start of the PRUNING phase, when processing the first $Y$ batch, we do not have a pruning threshold to initiate the pruning process. As a result, we must perform an additional full GEMM on the first $Y$ batch and implement our two-phase approach only on the subsequent $Y$ batches. To address this, we pick the centroid $y$ assigned to $x$ from the previous iteration and compute the distance between them (Algorithm~\ref{algo:pseudocode} Line 13). This results in a tight pruning threshold that we can use as soon as we start the \colorbox{softblue}{PRUNING} phase. This initial threshold is crucial, as it leads to discarding more than 95\% of the $y$ centroids based solely on the \colorbox{lightpeach}{GEMM} phase output. Thanks to the reliable pruning in ADSampling, this process does not introduce bias to keep the centroid assigned in the 1st iteration. 
It is worth noting that by recomputing the distance between $x$ and $y$, we use a distance that is valid for the current iteration.




\begin{table}[ht!]
\renewcommand{\tabcolsep}{3.0pt}
\centering
\caption{Hardware used for evaluation. All machines have 256GB of RAM and 32 threads, except Apple M4, which has 24GB of RAM and 14 threads.}
\vspace*{-4mm}
\resizebox{1\columnwidth}{!}{%
\begin{tabular}{llccccc}
\hline
\multirow{2}{*}{\textbf{Microarchitecture}} & \multirow{2}{*}{\textbf{CPU Model}} & \textbf{Freq.} & \textbf{L1} & \textbf{L2} & \textbf{L3} & \textbf{BLAS} \\
                                            &                                     & \textbf{GHz}   & \textbf{KiB} & \textbf{MiB} & \textbf{MiB} & \textbf{Library}                             \\ \hline
Intel Granite Rapids                     & Xeon 6975P-C                          & 3.9            & 48           & 2            & 480           & IntelMKL    \\
AMD Zen 5                                   & EPYC 9R45                           & 4.5            & 48           & 1            & 32           & AOCL-BLIS \\
AMD Zen 3                                   & EPYC 7R13                           & 3.6            & 32           & 0.5            & 16           & AOCL-BLIS \\
AWS Graviton 4                              & Neoverse V2                         & 2.8            & 64           & 2            & 36           & OpenBlas       \\
Apple M4 Pro  & M4 Pro                         & 4.5            & 128           & 16            & -           & Accelerate        \\
\hline
\end{tabular}
}
\label{tab:machine_details}
\vspace*{-3mm}
\end{table}
\begin{table}[t!]
\renewcommand{\tabcolsep}{3.0pt}
\centering
\caption{Vector embedding datasets used for evaluation}
\vspace*{-4mm}
\label{tab:datasets}
\resizebox{1.0\columnwidth}{!}{%
\begin{tabular}{lllccc}
\hline
\multicolumn{1}{l}{\textbf{Name}} & \multicolumn{1}{l}{\textbf{Embeddings}} & \textbf{Model} & \textbf{\# Vectors} & \textbf{Dim.} & \multicolumn{1}{c}{\textbf{Size (GB)$\uparrow$}} \\ \hline
Cohere    & Text  & EmbedV3-EN     & 10,000,000 & 1024 & 40.96 \\
arXiv     & Text  & InstructorXL   & 2,253,000  & 768  & 6.92  \\
OpenAI    & Text  & OpenAI         & 999,000    & 1536 & 6.14  \\
Wiki      & Text  & OpenAI         & 260,372    & 3072 & 3.20  \\
MXBAI     & Text  & MXBAI          & 769,382    & 1024 & 3.15  \\
Contriever& Text  & Contriever     & 999,000    & 768  & 3.07  \\
ImageNet  & Image & CLIP           & 1,281,167  & 512  & 2.62  \\
Yahoo     & Text  & MiniLM         & 677,305    & 384  & 1.04  \\
GloVe     & Word  & GloVe          & 1,183,514  & 200  & 0.95  \\
Yandex    & Text-to-Image & SE-ResNeXt & 1,000,000 & 200 & 0.80  \\
\hline

\end{tabular}
}
\vspace*{-3mm}
\end{table}


\section{Evaluation}\label{sec:eval}

We experimentally evaluate SuperKMeans on its speed and quality of clustering for vector search tasks across a range of modern vector embedding datasets. Additionally, we explore how the number of iterations and sampling impact the quality of the resulting centroids for vector search tasks. Section~\ref{sec:gpu} presents our evaluation on GPUs.

\subsection{Setup}

\vspace*{3mm}\noindent{\bf Hardware:} We primarily run experiments on an Intel Granite Rapids CPU with 256GB of RAM and 32 cores. In subsection~\ref{sec:eval:microarch}, we present an evaluation of SuperKMeans in four additional CPUs: AMD Zen 5, AMD Zen 3, AWS Graviton 4, and Apple M4 Pro. Table~\ref{tab:machine_details} presents detailed characteristics of these machines. Note that in our experiments, we use all the available cores and hardware optimizations (SIMD) for the corresponding CPU. 

\vspace*{3mm}\noindent{\bf Competitors:} We evaluated SuperKMeans against two SOTA $k$-means clustering libraries: FAISS, Scikit-Learn. We used FAISS v1.11.0 compiled to match the specific CPU characteristics and Scikit-Learn v1.8.0. To ensure a fair benchmarking process, we adjust the hyperparameters of each competitor and SuperKMeans (e.g., setting the same number of iterations and sampling rate for all competitors). In Scikit-Learn, we deactivated $k$-means++, in favor of random initialization. In Subsection~\ref{sec:eval:elkans}, we further justify our selection of competitors by showing that other libraries underperform when clustering vector embeddings. 

\vspace*{3mm}\noindent{\bf BLAS Libraries:}
We configured all competitors to use a BLAS library compiled for the respective CPU. In Table~\ref{tab:machine_details}, we detail which BLAS library was used in each CPU. It is noteworthy that the choice of BLAS implementation can significantly impact performance, as BLAS handles $\mathcolorbox{lightpeach}{X \boldsymbol{\cdot} Y}$--the main bottleneck of $k$-means. For instance, we found that OpenBLAS installed via \codeword{apt} on Linux can be up to 10x slower than when it is compiled from source with appropriate compilation flags for a specific CPU microarchitecture. 

\noindent{\bf Datasets:} We have selected ten vector embedding datasets for our evaluation, presented in Table~\ref{tab:datasets}. Most of these are commonly used to evaluate vector similarity search techniques~\cite{annbench, vibe, vectordbbench}. Additionally, we present an evaluation with classical vector datasets that do not stem from AI embedding models in subsection~\ref{sec:eval:classic}. 

\begin{figure*}[t!]
\centering
\includegraphics[width=1.0\linewidth]{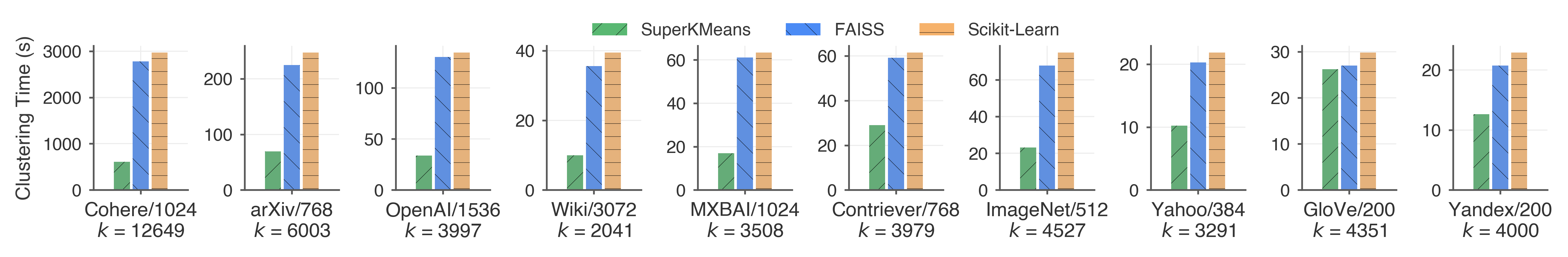}
\vspace*{-9.0mm}
\caption{Clustering performance (CPU) for all competitors with the number of clusters (k) set to $4 * \sqrt{N}$. SuperKMeans outperforms the competitors, particularly in high-dimensional datasets (d > 384), where it achieves 3-4x faster clustering. }
\vspace*{-4.0mm}
\label{fig:endtoend}
\end{figure*}
\begin{table}[t!]
\renewcommand{\tabcolsep}{1.5pt}
\centering
\caption{Quality of the generated centroids for VSS tasks. To measure recall, we do top-10 and top-100 IVF index search and explore 1\% of the total centroids ($k$). The quality of the centroids produced by SuperKMeans is on-par to the ones produced by FAISS, both in terms of recall and clusters' balance, while providing up to 4x faster build times.}
\vspace*{-4mm}
\label{tab:endtoend-recall}
\resizebox{1.0\columnwidth}{!}{%
\begin{tabular}{l l c c c c c c}
\hline
\multirow{2}{*}{\textbf{Dataset}} &
\multirow{2}{*}{\textbf{Algorithm}} &
\multirow{2}{*}{\textbf{$k$}} &
\textbf{Build} &
\textbf{Recall} &
\textbf{Recall} &
\textbf{Vectors} &
\textbf{WCSS} \\
 & & &
\textbf{Time (s)} &
\textbf{@100} &
\textbf{@10} &
\textbf{explored} &
\textbf{($\times 10^4$)} \\
\hline
\multirow{2}{*}{Cohere}
 & SuperKMeans & \multirow{2}{*}{12649} & \textbf{610.2} & 0.886 & 0.905 & 109721 & 508.22 \\
 & FAISS       &       & 2776.3        & 0.887 & 0.906 & 110834 & 506.47 \\
\hline
\multirow{2}{*}{arXiv}
 & SuperKMeans & \multirow{2}{*}{6003} & \textbf{69.55} & 0.931 & 0.969 & 24429 & 38.27 \\
 & FAISS       &      & 224.92        & 0.931 & 0.969 & 24435 & 38.25 \\
\hline
\multirow{2}{*}{OpenAI}
 & SuperKMeans & \multirow{2}{*}{3997} & \textbf{33.63} & 0.859 & 0.897 & 10795 & 62.14 \\
 & FAISS       &      & 130.40        & 0.860 & 0.895 & 10775 & 62.14 \\
\hline
\multirow{2}{*}{Wiki}
 & SuperKMeans & \multirow{2}{*}{2041} & \textbf{9.98} & 0.865 & 0.913 & 2875 & 16.07 \\
 & FAISS       &      & 35.62         & 0.866 & 0.917 & 2884 & 16.03 \\
\hline
\multirow{2}{*}{MXBAI}
 & SuperKMeans & \multirow{2}{*}{3508} & \textbf{16.93} & 0.923 & 0.964 & 8818 & 10087.9 \\
 & FAISS       &      & 61.12         & 0.925 & 0.966 & 8831 & 10089.3 \\
\hline
\multirow{2}{*}{Contriever}
 & SuperKMeans & \multirow{2}{*}{3979} & \textbf{29.03} & 0.919 & 0.977 & 11502 & 62.90 \\
 & FAISS       &      & 59.23         & 0.919 & 0.978 & 11446 & 62.90 \\
\hline
\multirow{2}{*}{ImageNet}
 & SuperKMeans & \multirow{2}{*}{4527} & \textbf{23.08} & 0.976 & 0.992 & 12982 & 29.16 \\
 & FAISS       &      & 67.70         & 0.975 & 0.990 & 13046 & 29.15 \\
\hline
\multirow{2}{*}{Yahoo}
 & SuperKMeans & \multirow{2}{*}{3291} & \textbf{10.22} & 0.886 & 0.931 & 6861 & 40.62 \\
 & FAISS       &      & 20.30         & 0.885 & 0.929 & 6798 & 40.62 \\
\hline
\multirow{2}{*}{GloVe}
 & SuperKMeans & \multirow{2}{*}{4351} & \textbf{26.21} & 0.722 & 0.813 & 13031 & 112.60 \\
 & FAISS       &      & 26.98         & 0.721 & 0.813 & 13111 & 112.62 \\
\hline
\multirow{2}{*}{Yandex}
 & SuperKMeans & \multirow{2}{*}{4000} & \textbf{12.61} & 0.879 & 0.922 & 9758 & 48.31 \\
 & FAISS       &      & 20.72         & 0.879 & 0.918 & 9500 & 48.34 \\
\hline
\end{tabular}
}
\vspace*{-4mm}
\end{table}

\begin{figure*}[t!]
\centering
\includegraphics[width=1.0\linewidth]{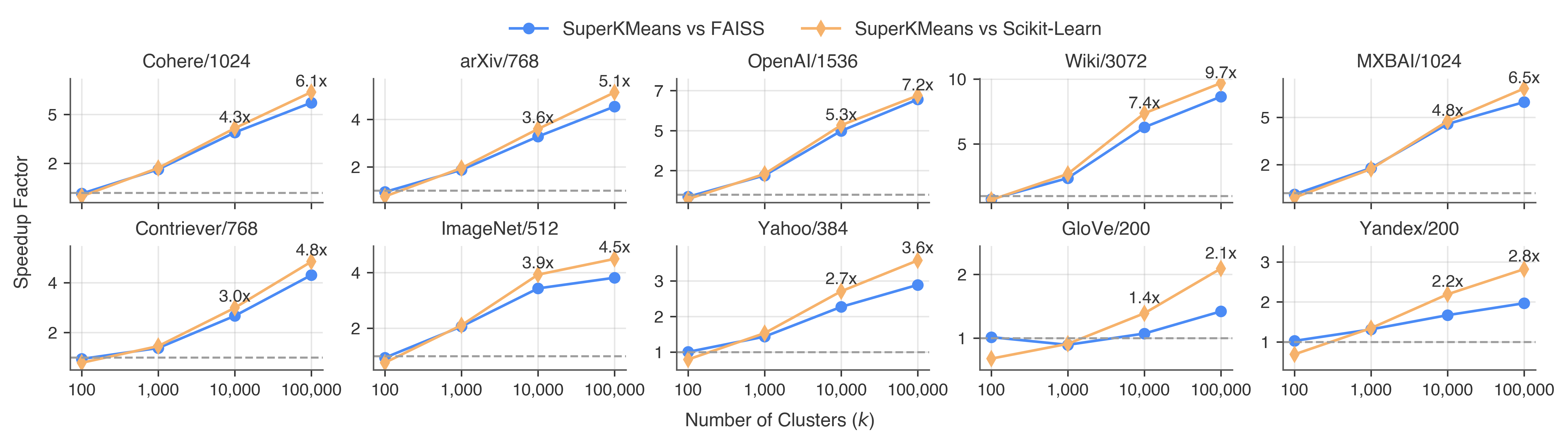}
\vspace*{-9.0mm}
\caption{Speedup of SuperKMeans (CPU) against all competitors at different values of $k$. SuperKMeans performs exceptionally well as the number of clusters increases. Conversely, for smaller values of $k$, SuperKMeans is on par with the competitors. 
}
\vspace*{-3.0mm}
\label{fig:varying-k}
\end{figure*}

\subsection{Clustering Time}\label{sec:eval:endtoend}
We evaluate the speed and quality of SuperKMeans against its competitors when clustering vector datasets. For this experiment, we fixed the number of iterations for all competitors at 25 and deactivated both sampling and early termination mechanisms. Furthermore, we set $k$ to $4* \sqrt{N}$, following the guidelines of many vector systems~\cite{faisscode, milvus}. As shown in Figure~\ref{fig:endtoend},  SuperKMeans achieves speedups of up to 4x on high-dimensional datasets, while providing substantial speedups on lower-dimensional datasets ($d \leq 384$) where pruning offers less benefits~\cite{pdx}. Furthermore, datasets with skewed distributions in their dimensions (OpenAI, Wiki) benefit more from pruning than datasets with normal distributions (Contriever, GloVe)~\cite{pdx, adsampling}. A higher intrinsic dimensionality also diminishes the benefits of pruning (e.g., GloVe).

SuperKMeans is not only faster but also equally accurate. Table~\ref{tab:endtoend-recall} shows the \textit{recall} obtained in VSS when using the clusters generated by SuperKMeans and FAISS as an IVF index. To measure recall, we search 1\% of the clusters to find the top-10 and top-100 nearest neighbors. SuperKMeans matches the optimal recall levels, with only minor deviations across datasets (< 0.005), which can be attributed to random seed variations and the loss in ADSampling's pruning. Note that both algorithms also yield a similar \textit{within-cluster sum of squares} value (WCSS, also known as \textit{inertia}). We have chosen not to include Scikit-Learn in Table~\ref{tab:endtoend-recall}, as its performance is similar to that of FAISS, both in runtime and index quality. 


\subsection{Scalability of $k$}\label{sec:eval:scalabilityk}
We evaluate each competitor's performance with different values of $k$, ranging from 100 to 100,000. In this experiment, we also fixed the number of iterations at 25 and deactivated the sampling and early termination mechanisms. Figure~\ref{fig:varying-k} shows that SuperKMeans scales better than its competitors as the number of clusters ($k$) increases, achieving up to 9x speedups on the Wiki dataset. However, it does not see benefits when $k$ is in the range of 10-100. This is because pruning finds less benefit when $Y$ is small and fits in the L1/L2 cache. However, it is important to note that these small values of $k$ are uncommon in VSS workloads. 

\begin{figure*}[t!]
\centering
\includegraphics[width=1.0\linewidth]{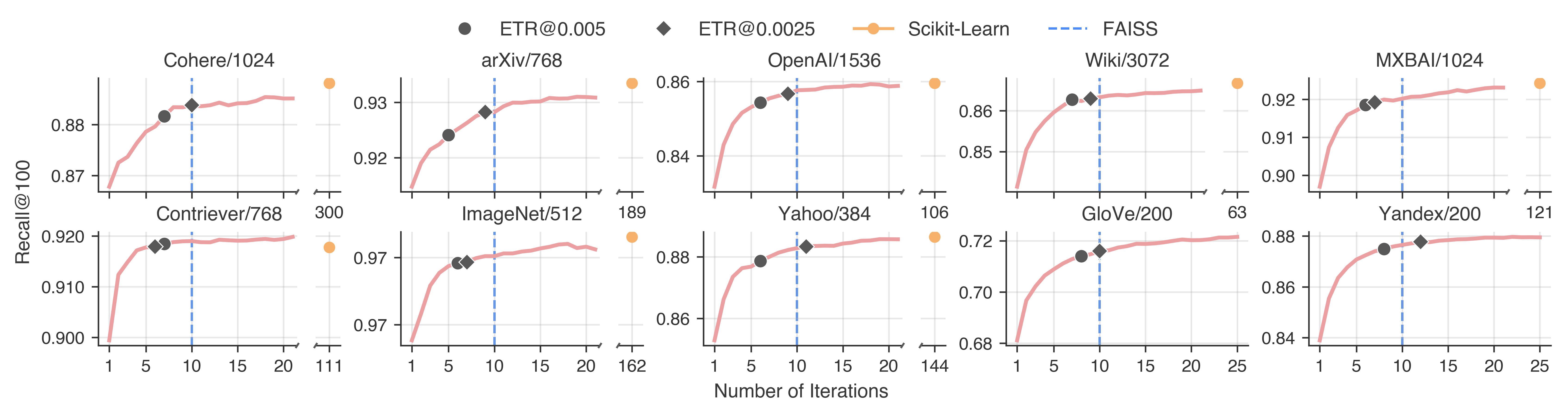}
\vspace*{-9.0mm}
\caption{The impact of early termination on VSS tasks. ETR generally results in earlier termination compared to FAISS, with minimal effect on recall. In contrast, the early termination in Scikit-Learn occurs much later, resulting in negligible improvements relative to the computational effort. Note: Scikit-Learn is not plotted in the datasets that need an angular distance, since this is an unsupported feature.}
\vspace*{-2.0mm}
\label{fig:etr}
\end{figure*}

\begin{figure*}[h]
\centering
\includegraphics[width=1.0\linewidth]{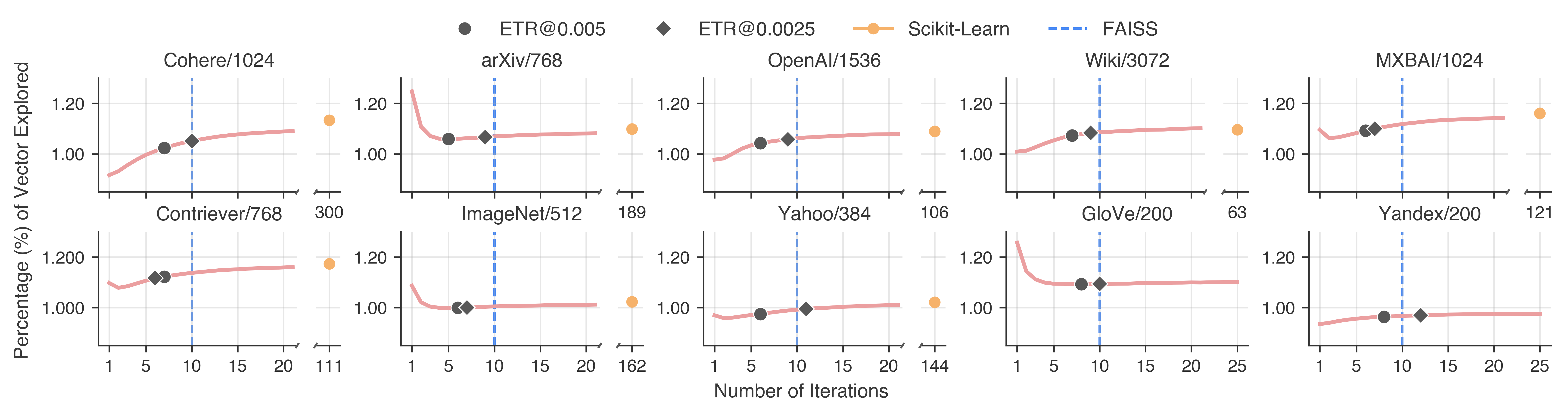}
\vspace*{-9.0mm}
\caption{In terms of the number of vectors explored during VSS, the datasets exhibit two distinct patterns as iterations progress: i) the number of vectors explored increases (Cohere, OpenAI) or ii) the number of vectors explored decreases (arXiv, ImageNet). These trends are effects of the cluster-balancing and refinement properties of our algorithm. However, the relative differences in the number of vectors explored across iterations are small.}
\vspace*{-2.0mm}
\label{fig:etr-explored}
\end{figure*}

\begin{figure}[h!]
\centering
\includegraphics[width=1.0\columnwidth]{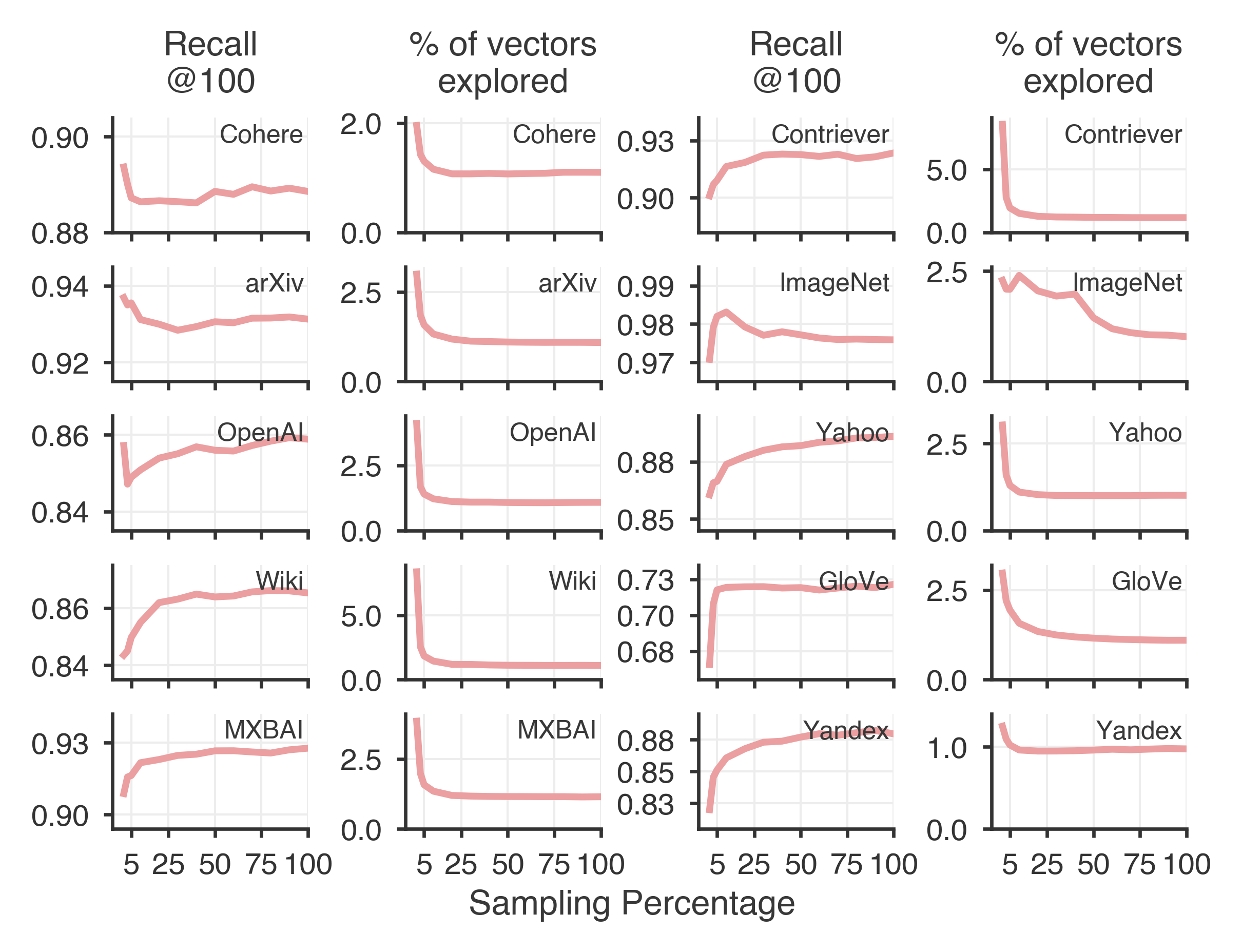}
\vspace*{-9.0mm}
\caption{Sampling has minimal impact on search quality (1st and 3rd columns). However, it results in less balanced clusters (2nd and 4th columns), thus increasing the number of vectors explored during VSS. Sampling 20-30\% of the data is enough to strike a balance between and balance. }
\vspace*{-2.0mm}
\label{fig:sampling}
\end{figure}

\subsection{Early Termination by Recall}\label{sec:eval:etr}
We evaluate the early-termination mechanism of each algorithm in terms of the quality of the resulting centroids and the number of iterations performed. In Scikit-Learn, we enable the default early termination based on centroid movement, keeping the default maximum number of iterations at 300. In FAISS, we enable the default early termination at 10 for IVF index construction. Finally, in SuperKMeans, we set a maximum number of iterations of 25 and configure ETR at two tolerance levels: 0.005 and 0.0025. The tolerance in ETR specifies how much improvement in the \textit{recall} metric must occur between two iterations for the algorithm to continue.

Figure~\ref{fig:etr} shows how recall improves across iterations, marking the points where early termination occurs. ETR results in earlier termination compared to FAISS, while having minimal impact on recall. In contrast, early termination in Scikit-Learn occurs much later, yielding minimal gains for the amount of computation done. Remarkably, just a single iteration is sufficient to achieve a relatively high retrieval quality; for example, the difference in recall between arXiv/768 at 1 and 20 iterations is only 0.015 recall points. 

However, as shown in Figure~\ref{fig:etr-explored}, the clusters tend to be unbalanced in the early iterations, potentially leading to increased work during VSS. ETR@0.005 effectively terminates the core loop at a sweet spot where both recall and cluster balance have stabilized. 
Figure~\ref{fig:etr-explored} shows two patterns as iterations progress: i) the number of vectors explored increases, and ii) the number of vectors explored decreases. We believe that, in the first scenario, clusters are being split in the region of the vector space that is not being queried. Conversely, in the second scenario, the cluster-splitting procedures reduce the size of clusters in the regions that are most frequently queried. In both cases, the quality of the search improves. 

\subsection{On the Effect of Sampling}\label{sec:eval:sampling}
Sampling is an effective technique for reducing clustering runtime by a factor proportional to the percentage of vectors sampled from the data. We are interested in understanding how sampling impacts the quality of the centroids for VSS tasks. We focus on a top-100 retrieval, exploring 1\% of the total number of clusters. The plots in the 1st and 3rd column of Figure~\ref{fig:sampling} show the recall achieved in VSS using the centroids produced by SuperKMeans under varying sampling fractions (from 1\% to 100\%). The plots in the 2nd and 4th column show the number of vectors explored within the most promising clusters. Interestingly, the recall for VSS shows only slight variation with higher sampling fractions (note the narrow y-axis range), suggesting that even with 10\% of the data, the centroids effectively represent the vector space. 

However, the percentage of vectors explored within those clusters decreases drastically with a larger sampling fraction, suggesting that the imbalance in cluster sizes is greater at smaller sample sizes. Note that the number of vectors explored affects the runtime of VSS--more vectors explored lead to more data accessed and more distance calculations. This effect is critical in out-of-distribution datasets (e.g., ImageNet), where the percentage of vectors explored dropped from 2\% with 5\% of the data to 1\% with 75\% of the data. However, in most embedding datasets, performing clustering with 20\%-30\% of the data points is enough to strike a balance between quality and clusters' balance. In some datasets (e.g., Cohere, OpenAI), recall spikes when sampling 1\% of the data. This is due to the imbalance in clusters at this sampling rate, which leads to the exploration of many more vectors, and thus an increase in recall.

Note that, with our specified values for $k$, FAISS' sampling will only trigger in the arXiv and Cohere datasets, where FAISS would sample 68\% and 32\% of the data, respectively. However, the results shown in Figure~\ref{fig:sampling} indicate that one can perform more aggressive sampling in most vector embedding datasets.

\subsection{Multi-threading Scalability}
\begin{figure}[t!]
\centering
\includegraphics[width=0.75\columnwidth]{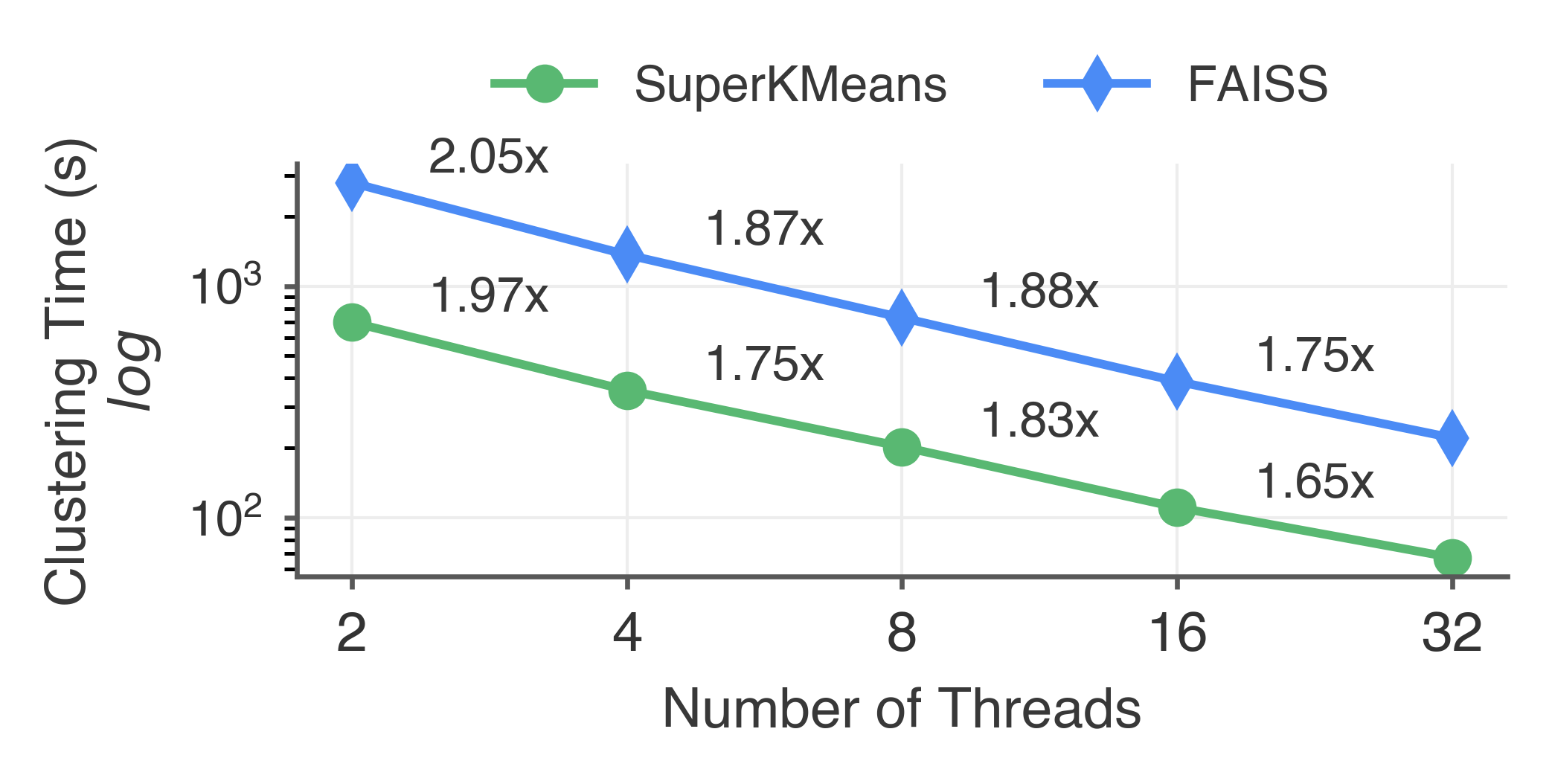}
\vspace*{-5.0mm}
\caption{Clustering time achieved with varying number of threads in the arXiv/768 dataset. Speedup factor between thread scaling is annotated. SuperKMeans scales almost linearly to the number of threads on the CPU.}
\vspace*{-2.0mm}
\label{fig:scalability}
\end{figure}

Figure~\ref{fig:scalability} shows the scalability of SuperKMeans with respect to the number of CPU threads (ranging from 2 to 32) on the arXiv/768 dataset. For this experiment, we deactivated the CPUs of our Intel Granite Rapids machine as needed. The results show that SuperKMeans scales nearly linearly with the number of available threads and delivers superior performance on machines with fewer cores. However, it does exhibit slightly worse scalability than FAISS. With 2 threads, SuperKMeans is 4x faster than FAISS, while with 32 threads, it is only 3.2x faster. Upon further examination, this is due to the uneven workload per thread during the \colorbox{softblue}{PRUNING} phase, where each $x''$ is handled by one thread. On average, we prune 97\% of centroids. However, this percentage can be even higher for certain vectors, leading some threads to perform more distance calculations than others. 

To address this uneven workload, we use \codeword{omp} \textit{dynamic scheduling}, allowing threads to request new chunks of work upon completing their current work. Here, it is crucial to strike a balance between the size of these chunks of work and the overhead associated with dynamic scheduling. In our scenario, a unit of work corresponds to a single vector $x''$. We found that a chunk of 8 vectors yields optimal performance. It is important to note that as the number of threads increases, dynamic scheduling starts behaving like \textit{static scheduling}. When we reach 128 threads, the dynamic scheduling of 8 units of work becomes equivalent to static scheduling, given that our $Y$ chunk size is set to 1024 vectors. While a custom thread-scheduling strategy could address this performance gap, the additional engineering effort required may not justify the potential gains. 

\subsection{Profiling}\label{sec:eval:microbench}
Table~\ref{tab:profiling} shows how the execution time is divided across the different phases of SuperKMeans. For this experiment, we set the number of iterations to 25, $k$ to $4 * \sqrt{N}$, and deactivated ETR. We highlight that the additional work of SuperKMeans (L2 identity, Random Rotation, Norms Calculation, and PDXification) represents little overhead compared to the core loop (\colorbox{lightpeach}{GEMM}, \colorbox{softblue}{PRUNING}, and 1st Iteration). Furthermore, it is remarkable that the 1st iteration (i.e., the full GEMM with all dimensions) takes around 15\% of the execution time of a 25-iteration SuperKMeans, demonstrating the effectiveness of SuperKMeans' two-stage approach to avoid work within the core loop. Finally, when ETR is activated, computing the ground truth of 1000 queries takes around 4-6\% of the runtime, and computing \textit{recall} in each iteration takes less than 1\% of the runtime. Thus, ETR incurs minimal overhead when considering its benefits.

\begin{table}[t!]
\renewcommand{\tabcolsep}{3.0pt}
\centering
\caption{Profiling of SuperKMeans on three datasets. The additional work of SuperKMeans (L2 identity, Random Rotation, Norms Calculation, and PDXification) represents minimal overhead compared to the total runtime. }
\vspace*{-4mm}
\label{tab:profiling}
\resizebox{1.0\columnwidth}{!}{%
\begin{tabular}{llll}
\toprule
\textbf{Stage} & \textbf{arXiv/768} & \textbf{Cohere/1024} & \textbf{OpenAI/1536} \\
\midrule
Core Loop (25 iters.)               & 62.3s (94.0\%) & 533.4s (96.6\%) & 26.6s (85.9\%) \\
- \colorbox{lightpeach}{GEMM}        & - 29.8s (44.9\%) & - 260.0s (47.1\%) & - 12.2s (39.4\%) \\
- \colorbox{softblue}{PRUNING}       & - 18.9s (28.5\%) & - 132.8s (24.1\%) & - 8.0s (25.7\%) \\
- 1st Iteration                     & - 8.8s (13.2\%) & - 96.9s (17.5\%)  & - 5.0s (16.1\%) \\
- L2 Identity                       & - 4.7s (7.2\%)  & - 42.4s (7.7\%)   & - 1.4s (4.6\%)  \\ \hline
Random Rotation                     & 1.0s (1.6\%)   & 5.0s (0.9\%)      & 1.9s (6.0\%)   \\ \hline
Update Centroids                    & 1.7s (2.6\%)   & 7.2s (1.3\%)      & 1.3s (4.3\%)   \\ \hline
Norms Calculation                   & 1.0s (1.6\%)   & 5.6s (1.0\%)      & 0.8s (2.7\%)   \\ \hline
PDXification                        & 0.1s (0.2\%)   & 0.7s (0.1\%)      & 0.3s (0.8\%)   \\ \hline
Centroids Splitting                 & $\approx$0 (0.0\%) & $\approx$0 (0.0\%) & $\approx$0 (0.0\%) \\ \hline
Centroids Initialization                    & $\approx$0 (0.1\%) & 0.1s (0.0\%)      & $\approx$0 (0.1\%) \\ \hline
Other                               & $\approx$0 (0.1\%) & 0.1s (0.0\%)      & 0.1s (0.2\%)   \\ \hline \hline
TOTAL                               & 66.2s          & 552.2s            & 31.0s          \\
\bottomrule
\end{tabular}
}
\vspace*{-0mm}
\end{table}

\subsection{Adjusting $d'$}\label{sec:eval:sweet}
There exists an optimal value for $d'$ that strikes an optimal balance between the work done by the \colorbox{lightpeach}{GEMM} phase and the number of vectors pruned during the \colorbox{softblue}{PRUNING} phase. Our empirical findings indicate that this balance is achieved when the percentage of vectors pruned in the \colorbox{softblue}{PRUNING} phase falls between 95\% and 97\%. Additionally, we found that, across most datasets, this level of pruning can be achieved by using around 12\% of $d$ in the \colorbox{lightpeach}{GEMM} phase. However, this is dataset-dependent. Moreover, as centroids become more stable, the efficiency of pruning improves, allowing us to reach our sweet spot with an even smaller $d'$.

To effectively manage this, we monitor the percentage of vectors pruned in each iteration and adjust $d'$ accordingly. If this percentage is $> 97\%$ (i.e., we are pruning too much), we reduce $d'$, to decrease the work of the \colorbox{lightpeach}{GEMM} phase. Conversely, if the percentage is $< 95\%$ (i.e., we are pruning too little), we increase $d'$, so that the \colorbox{lightpeach}{GEMM} phase performs more work. The goal is to maintain this percentage within the sweet range. For this, it is important to consider how aggressively we should adjust $d'$. Our empirical findings suggests that adjusting $d'$ by a factor of $ d' * 20\%$ leads to optimal performance. A smaller adjustment factor results in insufficient iterations for achieving the optimal $d'$ (recall that we are performing only a few iterations), while a larger factor can be too aggressive, causing $d'$ to fluctuate out of the sweet range across iterations. 

\begin{table}[t!]
\renewcommand{\tabcolsep}{3.0pt}
\centering
\caption{Ablation study of SuperKMeans key optimizations in the OpenAI/1536 dataset. Progressive pruning and a tighter initial threshold for the PRUNING phase contribute most to SuperKMeans's high performance. $\downarrow$ indicates slowdown.}
\vspace*{-4mm}
\label{tab:ablation}
\resizebox{1.0\columnwidth}{!}{%
\begin{tabular}{lcccc}
\hline
\textbf{Variant} & \textbf{\colorbox{lightpeach}{GEMM} (s)} & \textbf{\colorbox{softblue}{PRUNING} (s)} & \textbf{Other (s)} & \textbf{Total (s)} \\
\hline
Baseline (vanilla $k$-means) & 125.1 & - & - & 125.1 \\

\hline

SuperKMeans & 16.4 & 7.5 & 5.6 & 29.5 \\
- NO Initial threshold & 41.6 (2.5x $\downarrow$) & 8.3 (1.1x $\downarrow$) & 4.2 & 54.1 (1.83x $\downarrow$) \\

\hline

SuperKMeans & 16.4 & 7.5 & 5.6 & 29.5 \\
- NO PDXification & 16.4 & 8.8 (1.2x $\downarrow$) & 5.7 & 30.9 (1.05x $\downarrow$) \\
- NO Progressive pruning & 16.4 & 44.5 (6.0x $\downarrow$) & 5.5 & 66.4 (2.25x $\downarrow$) \\

\hline
SuperKMeans & 16.4 & 7.5 & 5.6 & 29.5 \\
- NO Adjustment of $d'$ & 22.5 (1.4x $\downarrow$) & 5.8 (1.3x $\uparrow$) & 5.3 & 33.6 (1.14x $\downarrow$) \\
\hline
\end{tabular}
}
\vspace*{-3mm}
\end{table}

\subsection{Ablation Study}\label{sec:eval:ablation}
Table~\ref{tab:ablation} presents an ablation study that examines the effect of four optimizations in our algorithm: i) Initial threshold for \colorbox{softblue}{PRUNING}, ii) progressive pruning (i.e., keep trying to prune the remaining vectors every 64 dimensions rather than computing the full distance), iii) PDXification, and iv) runtime adjustment of $d'$. In this study, we set the number of iterations to 25 and $k$ to 4 * $\sqrt{N}$. 

Progressive pruning is the main contributor to high performance. During progressive pruning, using the PDX layout (PDXification) results in a 1.2x speedup in the \colorbox{softblue}{PRUNING} phase compared to utilizing the row-major layout. In datasets where vectors are pruned at later dimensions, such as Contriever, disabling PDXification leads to a slowdown of 1.5x in the \colorbox{softblue}{PRUNING} phase. These findings are consistent with observations from~\cite{pdx, tribase}. The second most significant contributor to high performance is setting an initial threshold for the \colorbox{softblue}{PRUNING} phase. When this optimization is not utilized, a full GEMM is required with the first $Y$ batch to obtain a pruning threshold for $x$ in the current iteration, leading to a substantial slowdown. Lastly, adjusting $d'$ during runtime provides an additional 1.14x speedup, with a more pronounced effect in datasets of higher dimensionality.

\subsection{Classical Vector Datasets}\label{sec:eval:classic}
\begin{figure}[h!]
\centering
\includegraphics[width=0.70\columnwidth]{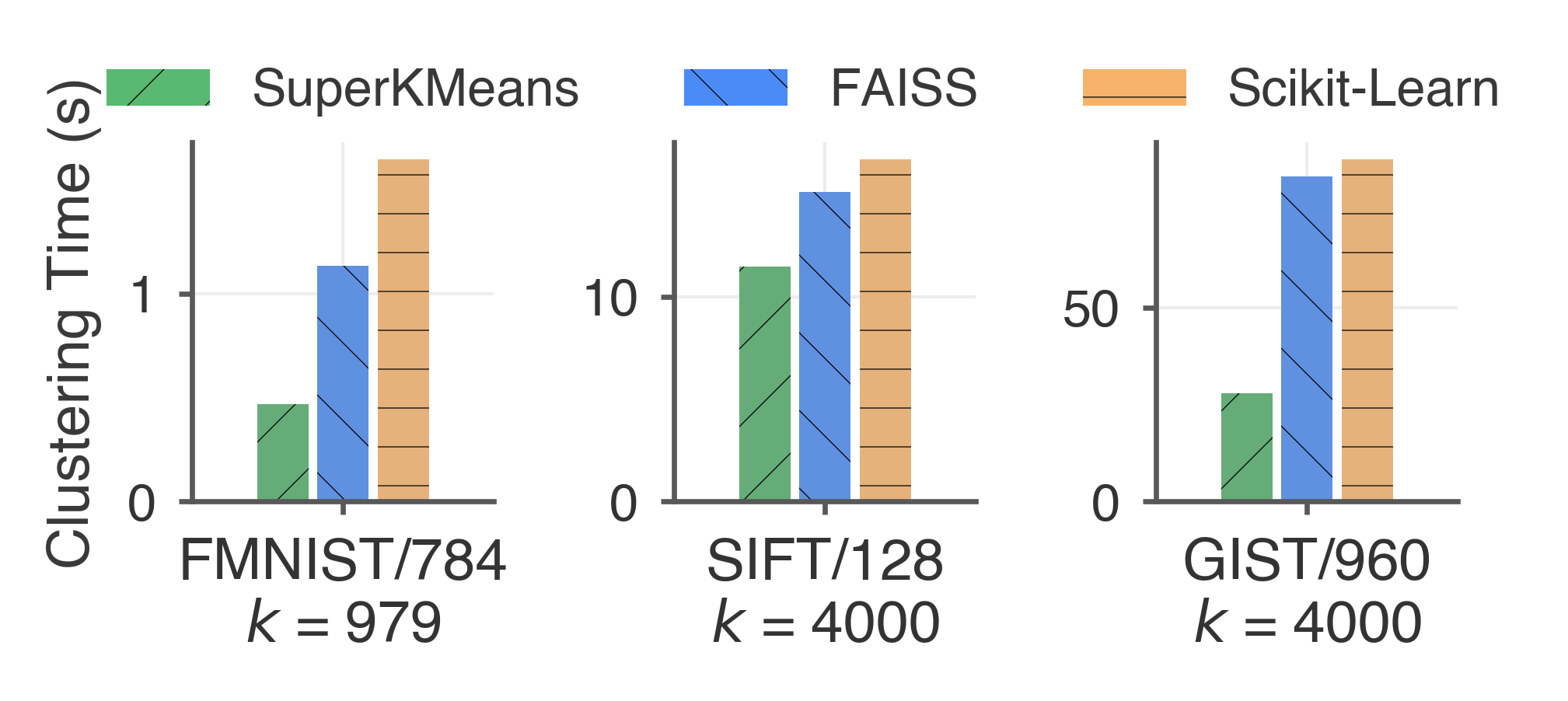}
\vspace*{-5.0mm}
\caption{SuperKMeans still outperforms the competitors in classic vector datasets that do not stem from AI embedding models, showing its portability across vector datasets. }
\vspace*{-4.0mm}
\label{fig:classic}
\end{figure}

We investigate how SuperKMeans performs in classical vector datasets that do not stem from AI embedding models. For this, we choose Fashion-MNIST ($N$=60K, $d$=784), SIFT ($N$=1M, $d$=128), and GIST ($N$=1M, $d$=960), which are all classical datasets used to evaluate VSS algorithms~\cite{annbench}. Figure~\ref{fig:classic} shows that SuperKMeans also accelerates $k$-means clustering in these datasets, further demonstrating its portability across scenarios.

\subsection{Across CPU Microarchitectures}\label{sec:eval:microarch}
\begin{figure}[t!]
\centering
\includegraphics[width=1.0\columnwidth]{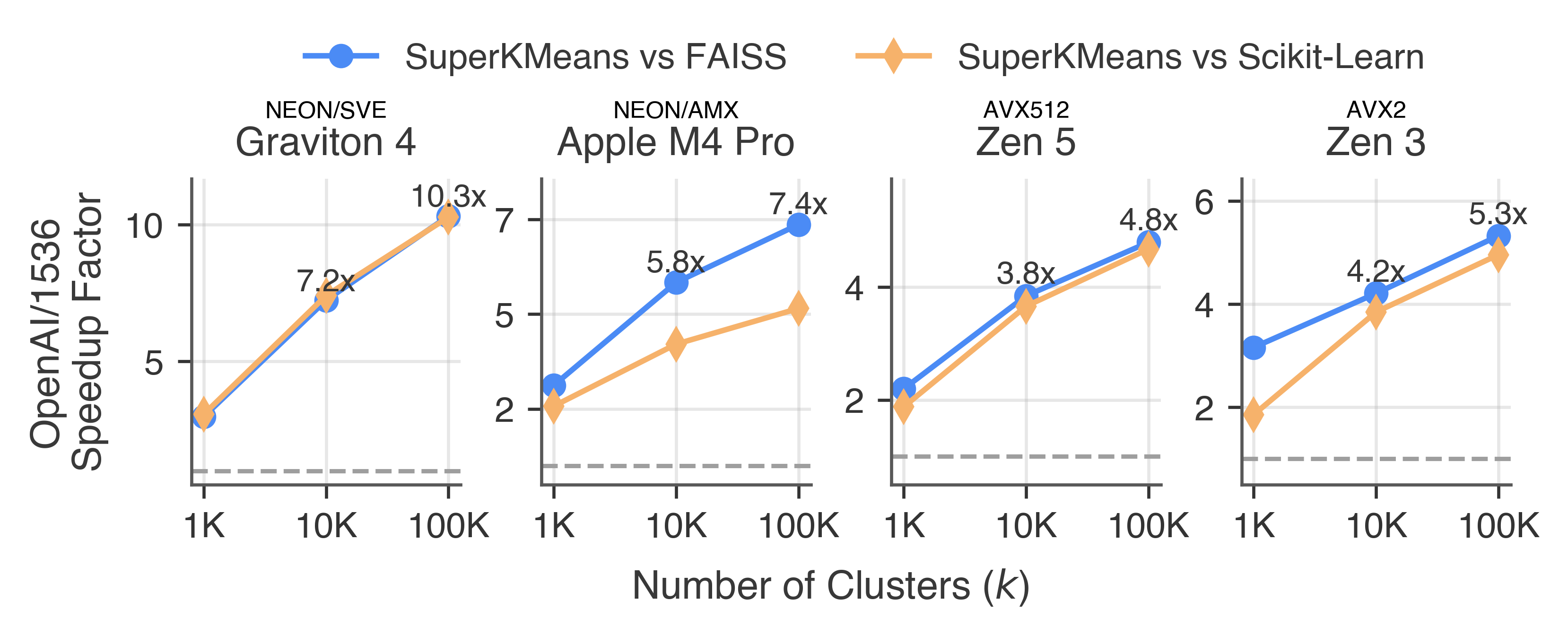}
\vspace*{-9.0mm}
\caption{SuperKMeans consistently defeats competitors, achieving speedups across all CPU microarchitectures}
\vspace*{-5.0mm}
\label{fig:microarch}
\end{figure}

We compare all competitors across the different CPU microarchitectures presented in Table~\ref{tab:machine_details}. We set the number of iterations to 25 and deactivated the sampling and early termination mechanisms. Figure~\ref{fig:microarch} shows the result of this experiment. SuperKMeans consistently defeats its competitors across all CPUs, demonstrating its portability across all SIMD instruction sets: AVX512 (Intel Granite Rapids, Zen5), AVX2 (Zen3), SVE/NEON (Graviton 4), and NEON+AMX (Apple M4 Pro).

SuperKMeans performs exceptionally well in the cost-effective Graviton 4, achieving speedups of up to 10x. This performance advantage can be attributed to the small register size of the Graviton 4 (128 bits), which makes the benefits of pruning more pronounced. FAISS heavily underperforms on the Apple M4 Pro machine, mainly due to the absence of NEON-based kernels across the FAISS codebase~\cite{kuffo2025bang}. In AMD's Zens, FAISS performs slightly worse than Scikit-Learn. On further inspection, we found that a larger batch size is beneficial on these AMD machines. Recall that SuperKMeans and FAISS process $X$ and $Y$ in batches of 4096 and 1024, respectively. By increasing the batch sizes to 40960 and 10240, FAISS achieved performance improvements of 10-30\% across datasets, at the expense of a larger intermediate buffer (1.56 GB). We believe that larger batch sizes are beneficial on Zen CPUs due to their higher L3 and DRAM memory bandwidth compared to Intel's.~\cite{kuffo2025bang}. However, larger batch sizes are not beneficial on the other CPUs. Conversely, SuperKMeans does not benefit from larger batches. This is because each $Y$ batch processed tightens the pruning threshold for the subsequent batch, resulting in less work overall. 

\begin{figure}[t!]
\centering
\includegraphics[width=0.98\columnwidth]{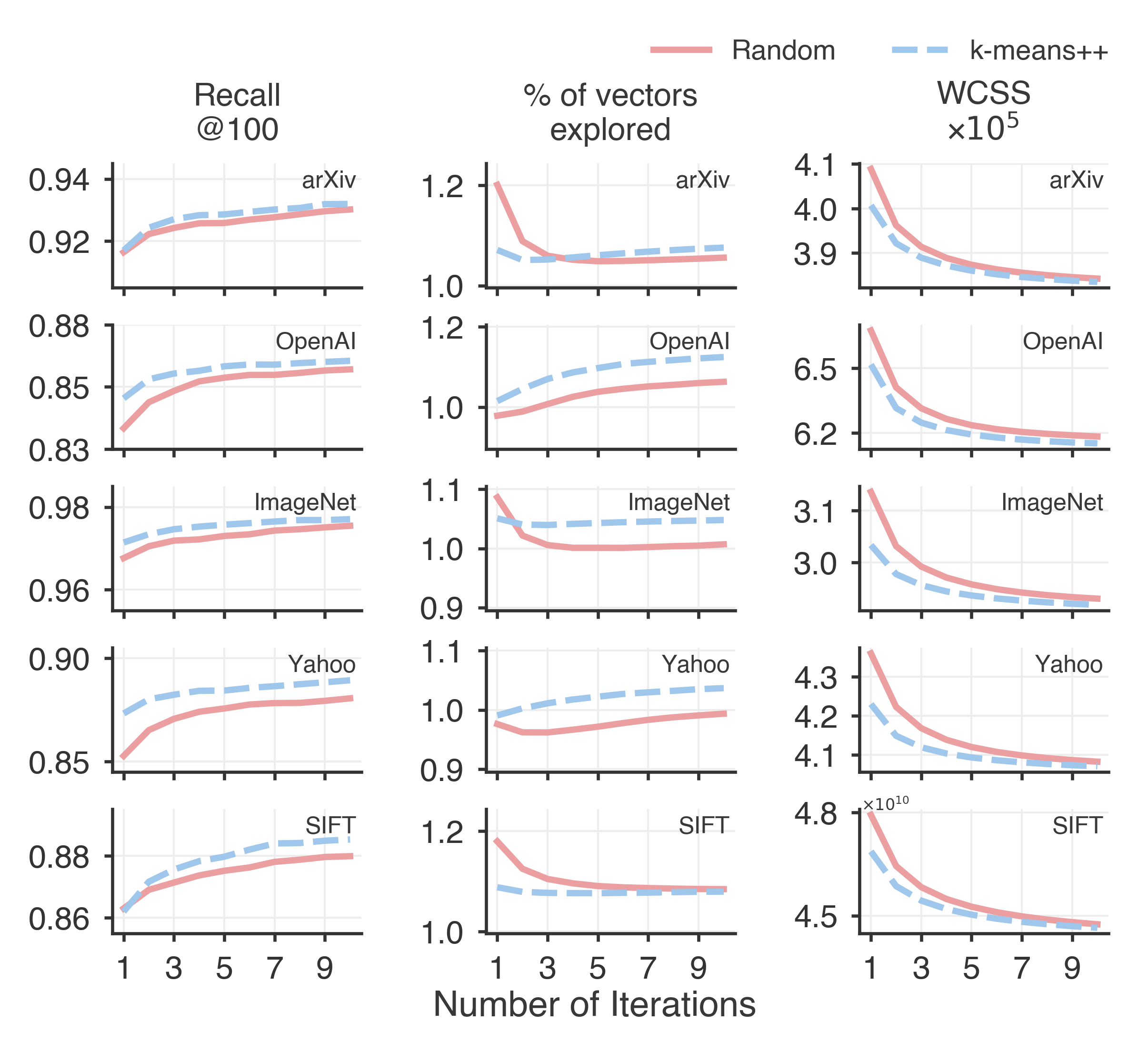}
\vspace*{-6.0mm}
\caption{The effect of initialization mechanisms on the resulting centroids for VSS tasks. $k$-means++ proves to be excessive, as its initialization is 10-100x slower. }
\vspace*{-2.0mm}
\label{fig:kmeans++}
\end{figure}

\subsection{Random vs k-means++ Initialization}\label{sec:eval:kmeans++}
We analyze the differences in the quality of centroids when using random versus $k$-means++ initialization in Scikit-Learn. Figure~\ref{fig:kmeans++} shows the results of this experiment, recording the quality of centroids across iterations in terms of their balance (2nd column) and recall (1st column) yielded in VSS tasks. Additionally, we show the WCSS in the 3rd column. The WCSS measures the compactness of clusters by summing the squared distances of each data point to its cluster's centroid. Notably, both initialization methods produce centroids of comparable quality in all aspects. However, $k$-means++ proves to be excessive, as its initialization process is significantly slower, taking 10-100x more time to complete the first iteration. 

An interesting observation is that $k$-means++ leads to exploring more vectors. This means that $k$-means++ yields slightly larger clusters, which slightly increases recall but slows down VSS. We believe this occurs because, in embedding datasets, most of the queried points lie in a dense region of the $d$-dimensional space~\cite{mohoney2024incremental, quake}. With random initialization, the sampling process selects a significant number of centroids from this region. In contrast, with $k$-means++, the first randomly selected centroid is likely to be within this region. However, the next chosen centroid will be located far from this region, since the algorithm chooses the point furthest away from the first centroid. Consequently, $k$-means++ prevents the algorithm from creating multiple centroids in this densely populated region, leading to the formation of larger clusters within it. Therefore, we argue that for VSS tasks, $k$-means++ is not only ineffective but can also be detrimental, potentially resulting in an imbalance among clusters within the region of the space that is queried the most.

\subsection{Space Complexity}\label{sec:eval:space}
SuperKMeans has a space complexity of $O(d(N+k))$, which is equivalent to that of the vanilla $k$-means algorithm. However, considering the hidden constants, SuperKMeans needs $2(N d) + k d + 3N + k$ space in memory. The first term $2(N d)$ comes from storing the vector collection $X$ twice: the original $X$ and the randomly rotated $X$. This duplication can be avoided by transforming $X$ in-place. In fact, Scikit-Learn provides a parameter named \codeword{copy_x} that, when set to \codeword{False}, enables in-place normalization of $X$. The second term $k d$, corresponds to the matrix of centroids $Y$. The term $3N$ corresponds to the space used by \codeword{assignments}, \codeword{distances}, and an additional buffer to store the partial norms of $X$. Finally, the term $k$ corresponds to the extra buffer to store the partial norms of $Y$. 


\subsection{Other k-means variants and libraries}\label{sec:eval:elkans}
\begin{figure}[h!]
\centering
\includegraphics[width=1.0\columnwidth]{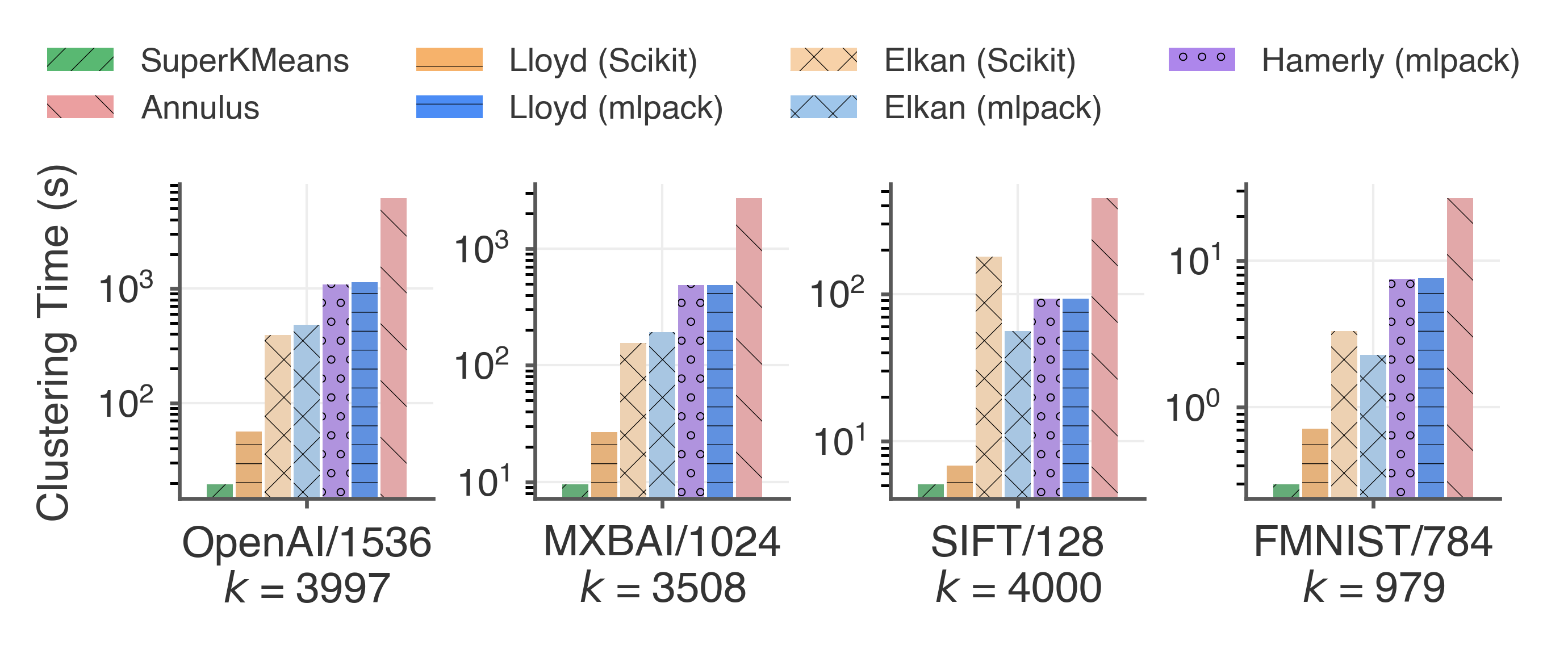}
\vspace*{-9.0mm}
\caption{Scikit-Learn's Lloyd's is faster than other $k$-means variants thanks to its use of GEMM routines. SuperKMeans is the best of both worlds: pruning distance computations while leveraging GEMM routines.  
}
\label{fig:elkan}
\end{figure}

Figure~\ref{fig:elkan} shows the clustering time of various $k$-means variants~\cite{hamerly2014accelerating, elkan2003using, zhakubayev2024using} compared to SuperKMeans. For this experiment, we used several implementations from Scikit-Learn~\cite{scikit}, mlpack~\cite{mlpack2023}, and the Fast $k$-means toolkit~\cite{kmeanstoolkit, zhakubayev2024using}. Notably, only Scikit-Learn's Lloyd's and SuperKMeans use GEMM routines. As a result, the other variants underperform, particularly when $k$ is large, as they are unable to take advantage of GEMM routines. Furthermore, many of these variants introduce a significant preprocessing step that acts as a major bottleneck, yielding minimal benefits since we are performing only a few iterations. We must stress that, with SuperKMeans, we have opened the door to an accelerated $k$-means that is CPU-friendly and can effectively leverage GEMM routines on specialized hardware, such as GPUs or the AMX unit in Apple Silicon. 

\begin{figure*}[t!]
\centering
\includegraphics[width=1.0\linewidth]{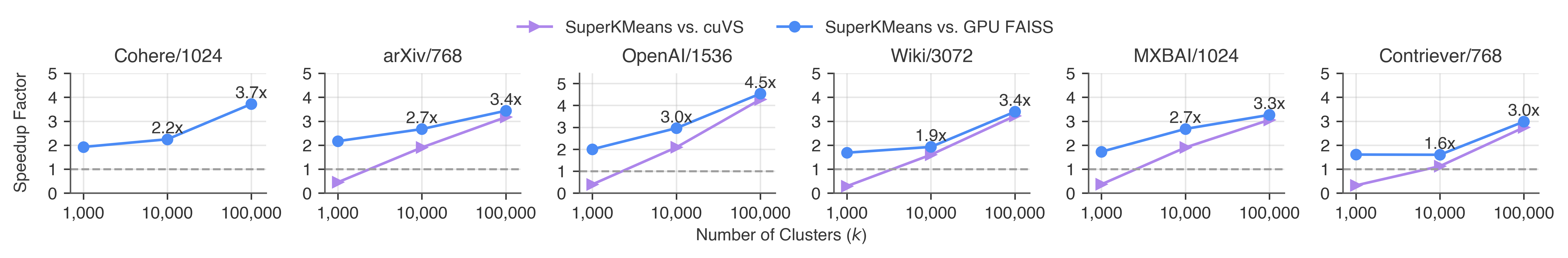}
\vspace*{-9.0mm}
\caption{Speedup of SuperKMeans against all competitors at different values of $k$ in NVIDIA H100 GPU. SuperKMeans performs exceptionally well as the number of clusters increases. Conversely, for smaller values of $k$, SuperKMeans loses to cuVS due to the preprocessing steps of SuperKMeans not being offloaded to the GPU. Note: cuVS crashes in the Cohere dataset.}
\vspace*{-2.0mm}
\label{fig:gpuperf}
\end{figure*}

\section{SuperKMeans in GPU}\label{sec:gpu}

The construction of IVF indexes is an attractive workload to offload to a GPU, as the workload is not a latency-sensitive task like VSS. This workload also offers enough opportunity for parallelism to leverage the thousands of threads that exist on GPUs. Therefore, we implemented the SuperKMeans algorithm not only on the CPU, but also on the GPU. We will first describe how we implemented the algorithm on the GPU and then compare its performance with two state-of-the-art GPU $k$-means clustering implementations.

\subsection{Implementation}

The GPU implementation of SuperKMeans follows the same algorithm outlined in Section~\ref{sec:superkmeans}, with one notable modification: the vector data batch size ($X$) is increased by a factor of 16 (from 4096 to 65,536) due to smaller batches performing worse on GPUs. The matrix multiplication of the front $d'$ dimensions (\colorbox{lightpeach}{GEMM} phase) is performed using cuBLAS~\cite{cublas}, NVIDIA's highly optimized matrix multiplication library. The \colorbox{softblue}{PRUNING} phase is implemented in a custom CUDA kernel~\cite{cudaguide}. The preprocessing steps are done on the CPU. Four CUDA streams are created, each of which is statically assigned batches of vectors. For each batch of vectors ($X$), a stream iterates over all batches of centroids ($Y$). This allows streams to independently schedule their kernels on the GPU, enabling the GPU to overlap both kernels and CPU-to-GPU memory copies~\cite{cudastreams}. 

The \colorbox{softblue}{PRUNING} kernel maps each vector to a single warp. The blocks in the PDX layout have 64 dimensions. Thus, a single warp can load all dimensions of a vector's block with two memory-coalesced accesses. The warp then considers which centroids need to be compared to the vector, and will load the dimensions of those centroids as well. The warp keeps comparing and pruning centroids until all are pruned, or only one remains.

The most crucial aspect of the algorithm is managing workload divergence per warp, as some warps may need to compare their vector against more centroids than others. Putting multiple warps in a single thread block will cause the streaming multiprocessor to wait until the last warp is finished processing its vector before scheduling the next thread block, effectively reducing occupancy. Therefore, we set the thread block size to a single warp, allowing every warp to process vectors independently. However, this also limits the maximum theoretical occupancy to 50\%, as each thread block needs to use at least two warps to reach 100\% theoretical occupancy. This low theoretical occupancy can cause the GPU to stall more often, as fewer warps can issue instructions to the streaming multiprocessor. We counteract this by leveraging the fact that, with a theoretical occupancy of 50\% due to the thread block size, we can use up to 64 registers per thread instead of 32 without impacting the theoretical occupancy. We use these registers to increase instruction-level parallelism on memory reads by fetching the dimensions of up to eight centroids simultaneously. This allows us to overlap the latencies on those memory accesses and reduces the impact of low occupancy~\cite{volkov, volkov-benchmarking-gpus-to-tune-dense-linear-algebra}.

\subsection{Evaluation Setup}

\noindent{\bf Hard- and software:} All GPU benchmarks were conducted on a system equipped with an AMD Zen 4 EPYC 9124 CPU and an NVIDIA H100 NVL GPU with 94 GB of HBM3 memory connected via PCI-Express 5.0×16. The software stack consisted of CUDA 13.1.0, NVIDIA driver version 575.57.08, cuBLAS 13.1.0, and GCC 13.3. All GPU benchmarks were executed inside a Docker container to ensure environmental isolation and reproducibility. 

\vspace*{3mm}\noindent{\bf Competitors:} We evaluated the GPU implementation of SuperKMeans against the GPU $k$-means clustering implementation in FAISS~\cite{gpufaisspaper} and cuVS~\cite{cuvs}. Notably, FAISS provides a cuVS-based IVF indexing implementation via its API. However, when using this interface, cuVS defaults to \textit{hierarchical} $k$-\textit{means}~\cite{karypis2000comparison}. Hereby, to maintain the fairness of our evaluation, we used the cuVS standard $k$-means clustering API directly. In Section~\ref{sec:hierarchical}, we present our evaluation of hierarchical $k$-means. Remarkably, the competitors use different NVIDIA libraries to implement the matrix multiplication step in $k$-means. For instance, FAISS uses the matrix multiplication kernels provided by NVIDIA's cuBLAS library. On the other hand, cuVS uses NVIDIA's CUTLASS library~\cite{cutlass}, which allows developers to fuse their kernels with highly optimized matrix multiplication kernels. cuVS leverages this to create a kernel that fuses the matrix multiplication and the nearest centroid search.

\vspace*{3mm}
\noindent{\bf Datasets:} We selected the six largest datasets from Table~\ref{tab:datasets}, as these heavy workloads are the most attractive for GPU acceleration.

\subsection{Evaluation Results}

We evaluate each competitor's performance with different numer of clusters ($k$), ranging from 1,000 to 100,000. For this experiment, we fixed the number of iterations to 25 and deactivated the sampling and early termination mechanisms. As shown in Figure~\ref{fig:gpuperf}, the GPU implementation of SuperKMeans achieves similar speedups as the CPU implementation over its competitors. SuperKMeans performs exceptionally well as $k$ increases, beating both FAISS and cuVS on all datasets when $k \geq 10,000$, with speedups of up to 3x at $k$=10,000. Remarkably, SuperKMeans always beats GPU FAISS and is only beaten by cuVS when $k$ is low. This is because SuperKMeans still performs preprocessing on the CPU, which incurs a constant cost independent of the number of centroids. When $k$ is larger, this fixed cost is amortized across longer iterations. However, for smaller values of $k$, this cost hurts the relative performance of GPU SuperKMeans. In all cases, cuVS is faster than GPU FAISS, but the performance converges with larger $k$. Note that cuVS crashes due to a memory bug when processing Cohere, the largest dataset.

Finally, Table~\ref{tab:cpu-gpu} compares the performance of GPU SuperKMeans and GPU FAISS with their CPU versions, demonstrating the advantages of offloading $k$-means to the GPU. cuVS is not included as it has no equivalent CPU implementation. The speedups of the GPU implementation of SuperKMeans are lower than those of FAISS because the preprocessing and tuning of $d'$ between iterations are still performed on the CPU in SuperKMeans. 

\begin{table}[t!]
\renewcommand{\tabcolsep}{7.0pt}
\centering
\caption{Comparison of clustering time (s) on CPU and GPU}
\vspace*{-4mm}
\label{tab:cpu-gpu}
\resizebox{1.0\columnwidth}{!}{%
\begin{tabular}{l|c|ccc|ccc}
\hline
\multirow{2}{*}{\textbf{Dataset}} 
& \multirow{2}{*}{\textbf{$k$}}
& \multicolumn{3}{c|}{\textbf{SuperKMeans}} 
& \multicolumn{3}{c}{\textbf{FAISS}} \\ 
& 
& \multicolumn{1}{c}{\textbf{CPU}} 
& \multicolumn{1}{c}{\textbf{GPU}} 
& \textbf{Speedup} 
& \multicolumn{1}{c}{\textbf{CPU}} 
& \multicolumn{1}{c}{\textbf{GPU}} 
& \textbf{Speedup} \\ 
\hline
Cohere     
& \multirow{6}{*}{10,000}
& \multicolumn{1}{c}{747.2} & \multicolumn{1}{c}{116.5} & 6.4$\times$  
& \multicolumn{1}{c}{2459.5} & \multicolumn{1}{c}{261.9} & 9.4$\times$ \\ 

arXiv      
& 
& \multicolumn{1}{c}{144.0} & \multicolumn{1}{c}{17.0}  & 8.4$\times$  
& \multicolumn{1}{c}{345.7}  & \multicolumn{1}{c}{45.6}  & 7.6$\times$ \\ 

OpenAI     
& 
& \multicolumn{1}{c}{74.7}  & \multicolumn{1}{c}{13.4}  & 5.6$\times$  
& \multicolumn{1}{c}{301.6}  & \multicolumn{1}{c}{39.9}  & 7.6$\times$ \\ 

Wiki       
& 
& \multicolumn{1}{c}{33.8}  & \multicolumn{1}{c}{9.0}   & 3.8$\times$  
& \multicolumn{1}{c}{194.0}  & \multicolumn{1}{c}{17.3}  & 11.2$\times$ \\ 

MXBAI      
& 
& \multicolumn{1}{c}{53.9}  & \multicolumn{1}{c}{7.7}   & 7.0$\times$  
& \multicolumn{1}{c}{188.8}  & \multicolumn{1}{c}{20.6}  & 9.1$\times$ \\

Contriever 
& 
& \multicolumn{1}{c}{71.0}  & \multicolumn{1}{c}{12.7}  & 5.6$\times$  
& \multicolumn{1}{c}{171.8}  & \multicolumn{1}{c}{20.3}  & 8.4$\times$ \\ 
\hline 
\end{tabular}
}
\vspace*{-2mm}
\end{table}


\begin{table}[t!]
\renewcommand{\tabcolsep}{1.5pt}
\centering
\caption{Quality of the generated centroids for VSS tasks between Vanilla and Hierarchical SuperKMeans. 
The quality of the centroids produced by Hierarchical SuperKMeans is slightly inferior in terms of recall. However, it results in more balanced clusters. PPC stands for Points per Cluster.
}
\vspace*{-4mm}
\label{tab:hierarchical-recall}
\resizebox{1.0\columnwidth}{!}{%
\begin{tabular}{l l c c c c c c c}
\hline
\multirow{2}{*}{\textbf{Dataset}} &
\multirow{2}{*}{\textbf{Algorithm}} &
\multirow{2}{*}{\textbf{$k$}} &
\textbf{Build} &
\textbf{Recall} &
\textbf{Vectors} &
\textbf{WCSS} &
\textbf{PPC} &
\textbf{PPC} \\
 & & &
\textbf{Time (s)} &
\textbf{@100} &
\textbf{explored (\%)} &
\textbf{($\times 10^5$)} &
\textbf{Mean} & \textbf{Std. Dev.} \\
\hline
\multirow{2}{*}{ImageNet}
 & Vanilla & \multirow{2}{*}{4527} & 10.50 & \textbf{0.975} & 1.01 & 2.93 & \multirow{2}{*}{283.0} & \textbf{110.5} \\
 & Hierarchical &  & \textbf{1.68} & 0.969 & \textbf{0.96} & 3.01 &  & 119.7 \\
\hline
\multirow{2}{*}{MXBAI}
 & Vanilla & \multirow{2}{*}{3508} & 8.12 & \textbf{0.920} & 1.12 & 1014.68 & \multirow{2}{*}{219.3} & 97.1 \\
 & Hierarchical &  & \textbf{1.84} & 0.903 & \textbf{1.01} & 1047.21 &  & \textbf{91.9} \\
\hline
\multirow{2}{*}{Wiki}
 & Vanilla & \multirow{2}{*}{2041} & 5.76 & \textbf{0.863} & 1.09 & 1.61 & \multirow{2}{*}{127.6} & 69.9 \\
 & Hierarchical &  & \textbf{2.93} & 0.844 & \textbf{0.96} & 1.64 &  & \textbf{63.1} \\
\hline
\multirow{2}{*}{Contriever}
 & Vanilla & \multirow{2}{*}{3979} & 13.84 & \textbf{0.919} & 1.14 & 6.31 & \multirow{2}{*}{248.8} & 114.9 \\
 & Hierarchical &  & \textbf{1.84} & 0.889 & \textbf{0.96} & 6.44 &  & \textbf{96.7} \\
\hline
\multirow{2}{*}{OpenAI}
 & Vanilla & \multirow{2}{*}{3997} & 16.90 & \textbf{0.858} & 1.06 & 6.23 & \multirow{2}{*}{249.9} & 119.0 \\
 & Hierarchical &  & \textbf{4.19} & 0.836 & \textbf{0.91} & 6.39 &  & \textbf{117.3} \\
\hline
\multirow{2}{*}{arXiv}
 & Vanilla & \multirow{2}{*}{6003} & 31.65 & \textbf{0.928} & 1.07 & 3.84 & \multirow{2}{*}{375.3} & 140.3 \\
 & Hierarchical &  & \textbf{4.44} & 0.910 & \textbf{0.98} & 3.94 &  & \textbf{136.7} \\
\hline
\multirow{2}{*}{Cohere}
 & Vanilla & \multirow{2}{*}{12649} & 297.66 & \textbf{0.884} & 1.05 & 51.10 & \multirow{2}{*}{790.6} & 353.6 \\
 & Hierarchical &  & \textbf{24.95} & 0.854 & \textbf{0.88} & 52.62 &  & \textbf{341.2} \\
\hline
\end{tabular}
}
\vspace*{-4mm}
\end{table}

\section{Hierarchical SuperKMeans}\label{sec:hierarchical}

Hierarchical $k$-means~\cite{karypis2000comparison} has recently emerged as an alternative for clustering large collection of vector embeddings~\cite{micronn, kmeanselastic, turbopuffer, cuvs, jin2026curator, spfresh}. This variant is appealing because it reduces the computational complexity of the vanilla $k$-means algorithm by dividing the clustering process into two phases: MESO-CLUSTERING and FINE-CLUSTERING. In the MESO-CLUSTERING phase, a coarser clustering is performed by creating only $\sqrt{k}$ clusters, referred to as meso-clusters. Subsequently, during the FINE-CLUSTERING phase, $k$-means is run within each meso-cluster built in the first phase, with the number of clusters set to $\sqrt{n_{i}}$, which corresponds to the number of points inside that i-th meso-cluster. This approach effectively reduces the computational complexity from $O(N * k * d)$ to $O(N * \sqrt{k} * d)$. Throughout both phases, we leverage SuperKMeans to accelerate the pairwise distance calculations, addressing the primary bottleneck of the algorithm.

\begin{figure}[t!]
\centering
\includegraphics[width=1.0\columnwidth]{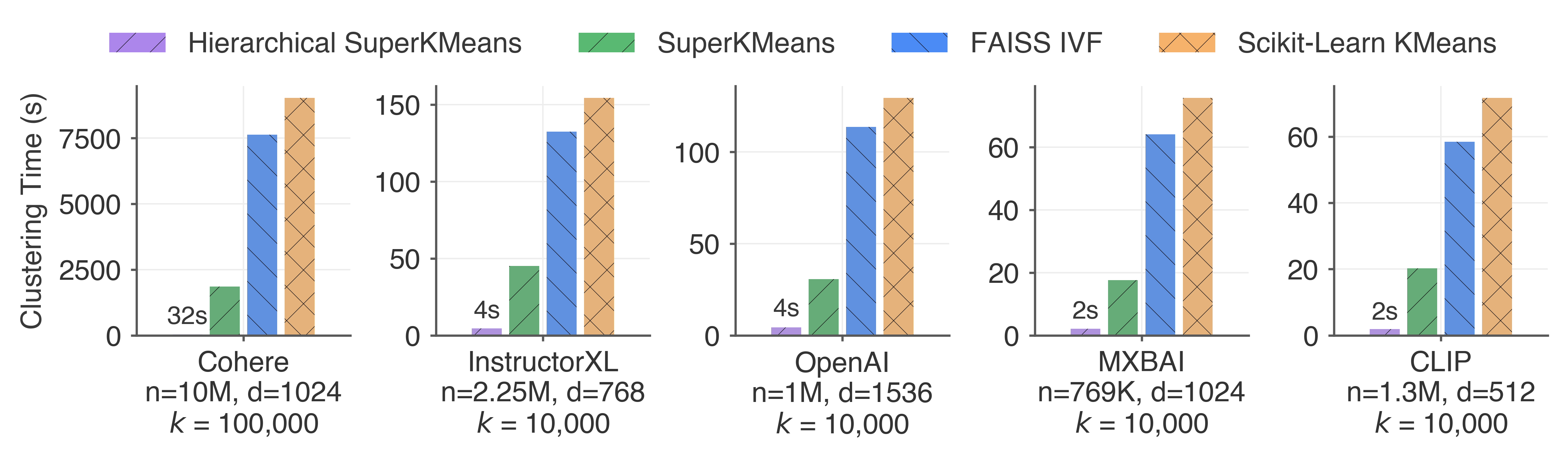}
\vspace*{-8.0mm}
\caption{Hierarchical SuperKMeans achieves unprecedent performance, achieving 250x faster clustering than FAISS.}
\vspace*{-2.0mm}
\label{fig:hierarchical_clustering}
\end{figure}

\subsection{Evaluation of Hierarchical SuperKMeans}

We evaluate Hierarchical SuperKMeans against Vanilla SuperKMeans, FAISS, and Scikit-Learn. Our experimental framework remains the same as in Section~\ref{sec:eval}, but using only 10 iterations. For Hierarchical SuperKMeans, we set the number of iterations to 3 in the MESO-CLUSTERING phase and 5 in the FINE-CLUSTERING phase. As shown in Figure~\ref{fig:hierarchical_clustering}, Hierarchical SuperKMeans achieves remarkable performance, being able to compute the centroids for the Cohere dataset 250x times faster than FAISS and 60x faster than Vanilla SuperKMeans. Furthermore, both SuperKmeans and Hierarchical SuperKMeans complete the final assignments 4x faster than FAISS, resulting in a 30x faster end-to-end runtime. These results demonstrate that SuperKMeans' pruning is versatile, as it accelerates the core $k$-means primitive. 

Nevertheless, hierarchical $k$-means alters the achieved recall and cluster balance because it fundamentally changes how clusters are built. Table~\ref{tab:hierarchical-recall} compares vanilla and hierarchical $k$-means in terms of the recall and balance of the produced centroids. Hierarchical SuperKMeans degrades recall up to 3\%. However, it significantly improves cluster balance. For instance, in the Cohere dataset, the hierarchical $k$-means clusters lead to exploring 16\% fewer vectors. However, when increasing the number of clusters explored in hierarchical $k$-means to match the number of vectors explored in vanilla $k$-means, we still observe a gap in recall of up to 1\%. The latter shows that there exists a gap in the achievable retrieval quality between hierarchical and vanilla $k$-means. 

\vspace*{3mm}\noindent{\bf On the effect of sampling and number of iterations}: Regarding sampling, we observed a similar behavior to Vanilla SuperKMeans: 20-30\% sampling is sufficient to achieve optimal retrieval quality. For the number of iterations, we performed a grid search to identify the Pareto optimal number of iterations for each of the two phases across datasets. We found that 3 iterations for MESO-CLUSTERING and 5 for FINE-CLUSTERING are the best configuration across datasets to achieve optimal quality.


\section{Discussion}\label{sec:discussion}
SuperKMeans is a significant advancement for $k$-means algorithms, particularly for high-dimensional vector embeddings. Notably, our approach accelerates the core $k$-means primitive. Thus, it can be used for other $k$-means variants, such as hierarchical $k$-means~\cite{karypis2000comparison, jin2026curator, kmeanselastic} and MiniBatchKMeans~\cite{micronn, jin2006fast}. Furthermore, SuperKMeans' scalability to larger values of $k$ makes it especially useful in large-scale settings where billions of vectors must be clustered. We must stress that not only can IVF indexes benefit from SuperKMeans, but any indexing pipeline that requires a coarse index before applying other techniques, such as Product Quantization~\cite{pqivf} or fine-grained indexing/quantization per cluster~\cite{spann, rabitq, rabitqext}. Similarly, Early Termination by Recall (ETR) can be plugged into any indexing pipeline that performs iterative refinement. Notably, the idea behind ETR was inspired by early termination strategies for VSS queries~\cite{darth, quake}. 

Weaviate has introduced TileEncoder as an alternative to $k$-means clustering~\cite{weaviatetileencoder}. TileEncoder divides the vector space into tiles that reflect the dimensions distribution and assigns vectors to tiles based on their values, leading to significantly faster clustering times. However, TileEncoder has a notable impact on retrieval quality~\cite{weaviatetileencoder}, and is mainly used for Product Quantization, where clustering is performed on a limited number of dimensions at a time (typically 4 to 16). On the other hand, CrackIVF proposes avoiding upfront clustering costs by progressively building an IVF index while answering VSS queries, using brute-force search initially~\cite{crackivf}. CrackIVF is especially useful for querying cold data with sparse access patterns where the cost of upfront indexing might not be justified. SuperKMeans, in this context, complements CrackIVF by reducing the upfront indexing cost when indexing is needed.

\vspace*{3mm}
\noindent{\bf Vanilla vs Hierarchical $k$-means:} Hierarchical SuperKMeans drastically reduces construction time, albeit at a slightly lower recall in VSS tasks. We believe this trade-off in recall is a necessary price to pay when indexing large collections (+10M points), which would otherwise take hours with SuperKMeans or days with FAISS. A technique used by cuVS to close this gap is to perform a third phase called REFINEMENT-CLUSTERING. This phase corresponds to a couple of vanilla $k$-means iterations after the first two phases. However, these iterations can take hours or days on CPUs, depending on the dataset size. We can speed up these refinement iterations with SuperKMeans; however, we argue that performing them defeats the initial purpose of using hierarchical $k$-means: to achieve blazing-fast clustering in large collections. Nevertheless, future work should focus on addressing this gap more efficiently. 

\section{Conclusions and Future Work}\label{sec:conclusion}

We have presented and evaluated SuperKMeans, a variant of $k$-means that accelerates the clustering of high-dimensional vectors by \textit{efficiently} avoiding unnecessary distance calculations during the assignment step of $k$-means. SuperKMeans' novelty lies in carefully interleaving GEMM routines and pruning kernels, a feat that previous $k$-means variants have not achieved. 
SuperKMeans achieves faster clustering speeds than competing methods in CPUs (up to 10x faster) and GPU (up to 4x faster), while maintaining the quality of the centroids for vector similarity search tasks. SuperKMeans performance improvement is particularly notable as the number of clusters ($k$) increases. 
Our study also provides insights on the behavior of $k$-means with high-dimensional embeddings. Our findings show that: i) just 5 to 10 iterations of $k$-means are enough to achieve optimal retrieval quality, ii) using only 20\% to 30\% of the data points is sufficient for generating high quality centroids, and iii) more complex initialization mechanisms, such as $k$-means++, do not result in a significant quality improvement on the generated centroids for retrieval tasks.  
As future work, we believe that SuperKMeans can be used to cluster datasets that have undergone a dimensionality-reduction process to speed up clustering~\cite{zhakubayev2022clustering}. Finally, we plan to improve upon our initial GPU implementation. For instance, incorporating Flash K-Means for efficient GPU IO~\cite{flashkmeans}. 

\balance
\bibliographystyle{ACM-Reference-Format}
\bibliography{_main}


\appendix



\end{document}